\documentclass[10pt,twocolumn,letterpaper]{article}

\usepackage{titling}
\usepackage{chngcntr}
\usepackage{cvpr}
\usepackage{times}
\usepackage{epsfig}
\usepackage{graphicx}
\usepackage{amsmath}
\usepackage{amssymb}

\usepackage{url}
\usepackage{xspace}
\usepackage{booktabs}
\usepackage{mathtools}
\usepackage{bbm}
\usepackage{enumitem}
\usepackage{stmaryrd}
\usepackage{xcolor}
\usepackage{algorithmicx,algorithm}
\usepackage{algpseudocode}
\usepackage{xcolor}
\usepackage{caption}
\usepackage{subcaption}
\usepackage{arydshln}
\usepackage{comment}
\usepackage{pifont}

\usepackage[pagebackref=true,breaklinks=true,letterpaper=true,colorlinks,bookmarks=false]{hyperref}

\renewcommand{\vec}[1]{\mathbf{#1}}
\newcommand{\noask}{\textsc{None}\xspace}
\newcommand{\randomask}{\textsc{Random}\xspace}
\newcommand{\askfirst}{\textsc{First}\xspace}
\newcommand{\learntoask}{\textsc{Learned}\xspace}
\newcommand{\oracleask}{\textsc{Teacher}\xspace}

\newcommand{\navpol}{\ensuremath{\pi_{\textrm{nav}}}\xspace}
\newcommand{\askpol}{\ensuremath{\pi_{\textrm{ask}}}\xspace}

\newcommand{\iiil}{\textsc{I3L}\xspace}
\newcommand{\asknav}{\textsc{AskNav}\xspace}
\newcommand{\testseen}{\textsc{Test seen}\xspace}
\newcommand{\testunseen}{\textsc{Test unseen}\xspace}

\newcommand{\cmark}{\ding{51}}
\newcommand{\xmark}{\ding{55}}

\DeclarePairedDelimiter\floor{\lfloor}{\rfloor}

\cvprfinalcopy % *** Uncomment this line for the final submission

\def\cvprPaperID{6028} % *** Enter the CVPR Paper ID here

\ifcvprfinal\fi
\begin{document}

%%%%%%%%% TITLE
\title{Vision-based Navigation with Language-based Assistance \\ via Imitation Learning with Indirect Intervention}

\author{Khanh Nguyen \\
University of Maryland, College Park\\
{\tt\small kxnguyen@cs.umd.edu}
% For a paper whose authors are all at the same institution,
% omit the following lines up until the closing ``}''.
% Additional authors and addresses can be added with ``\and'',
% just like the second author.
% To save space, use either the email address or home page, not both
\and
Debadeepta Dey, Chris Brockett, Bill Dolan\\
Microsoft Research, Redmond\\
{\tt\small \{dedey,Chris.Brockett,billdol\}@microsoft.com}
}

\maketitle

\begin{abstract}

We present Vision-based Navigation with Language-based Assistance (VNLA), a grounded vision-language task where an agent with visual perception is guided via language to find objects in photorealistic indoor environments.
 The task emulates a real-world scenario in that (a) the requester may not know how to navigate to the target objects and thus makes requests by only specifying high-level end-goals, and (b) the agent is capable of sensing when it is lost and querying an advisor, who is more qualified at the task, to obtain language subgoals to make progress. 
To model language-based assistance, we develop a general framework termed Imitation Learning with Indirect Intervention (I3L), and propose a solution that is effective on the VNLA task. 
Empirical results show that this approach significantly improves the success rate of the learning agent over other baselines in both seen and unseen environments. 

Our code and data are publicly available at \url{https://github.com/debadeepta/vnla}.

\end{abstract}

\section{Introduction}
\begin{figure*}[t!]
  \centering
  \includegraphics[width=.95\linewidth]{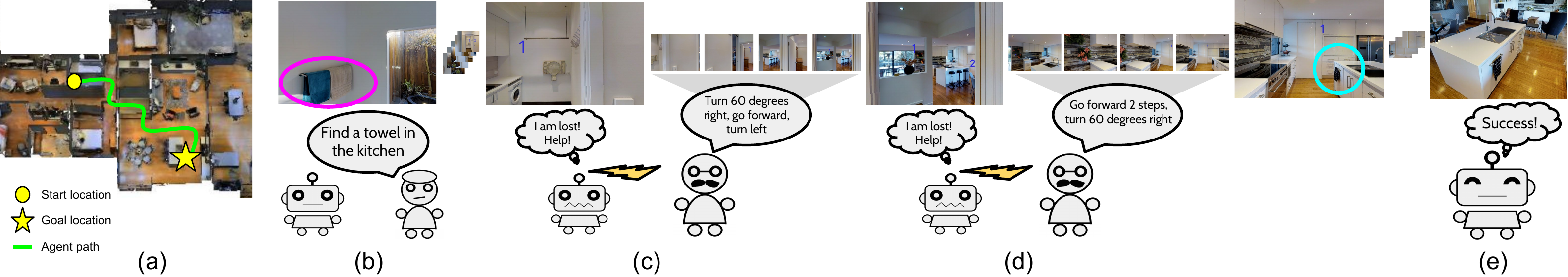}
  \setlength\abovecaptionskip{3pt}
  \caption{An example run in an unseen environment. (a) A bird-eye view of the environment annotated with the agent's path. The agent observes the environment only through a first-person view. (b) A requester (wearing a hat) asks the agent to ``find a towel in the kitchen". Two towels (pink circle) are in front of the agent but the room is labeled as a ``bathroom". The agent ignores them without being given the room label. (c) The agent escapes the bathroom but runs into an unfamiliar region. Sensing that it is lost, the agent signals the advisor (with mustache) for help. The advisor responds with an ``easier" low-level subgoal ``turn 60 degrees right, go forward, turn left". (d) After executing the subgoal, the agent is closer to the kitchen but is still confused. It thus requests help one more time. After making this request, the agent has exhausted its request budget and can only rely on its own. (e) Executing the second subgoal helps the agent see the target towel (cyan circle). It successfully walks to the goal without further assistance. A video demo is at \url{https://youtu.be/Vp6C29qTKQ0}.}
  \label{fig:problem}
\end{figure*}

Rich photorealistic simulators are finding increasing use as research testbeds and precursors to real-world embodied agents such as self-driving cars and drones \cite{shah2018airsim,dosovitskiy2017carla,kempka2016vizdoom}. 
Recently, growing interest in grounded visual navigation from natural language is facilitated by the development of more realistic and complex simulation environments \cite{anderson2018vision,kolve2017ai2,brodeur2017home,puig2018virtualhome,xiazamirhe2018gibsonenv,wu2018building,savva2017minos} in place of simple toy environments \cite{bisk2016natural,chen2011learning,macmahon2006walk,branavan2009reinforcement,branavan2012learning}. 
%Recently, \cite{anderson2018vision} propose a visual-language navigation task in which an agent must negotiate indoor environments according to crowd-sourced turn-by-turn instructions. 
Several variants of this task have been proposed.
In \cite{anderson2018vision}, agents learn to execute natural language instructions crowd-sourced from humans. 
\cite{das2018embodied} train agents to navigate to answer questions about objects in the environment.
\cite{de2018talk} present a scenario where a guide and a tourist chat to direct the tourist to a destination.

In this paper, we present Vision-based Navigation with Language-based Assistance (VNLA), a grounded vision-language task that models a practical scenario: a mobile agent, equipped with monocular visual perception, is requested via language to assist humans with finding objects in indoor environments.
A realistic setup of this scenario must (a) not assume the requester knows how to accomplish a task before requesting it, and (b) provide additional assistance to the agent when it is completely lost and can no longer make progress on a task. 
To accomplish (a), instead of using detailed step-by-step instructions, we request tasks through high-level instructions that only describe end-goals (e.g. ``Find a pillow in one of the bedrooms"). 
To fulfill (b), we introduce into the environment an advisor who is present at all times to assist the agent upon request with low-level language subgoals such as ``Go forward three steps, turn left".
In VNLA, therefore, the agent must (a) ground the object referred with the initial end-goal in raw visual inputs, (b) sense when it is lost and use an assigned budget for requesting help, and (c) execute language subgoals to make progress.

VNLA  motivates a novel imitation learning setting that we term Imitation Learning with Indirect Intervention (\iiil). 
In conventional Imitation Learning (IL), a learning agent learns to mimic a teacher, who is only available at training time, by querying the teacher's demonstrations on situations the agent encounters.
I3L extends this framework in two ways. 
First, an advisor is present in the environment to assist the agent not only during training time but also at test time. 
Second, the advisor assists the agent not by directly making decisions on the agent's behalf, but by modifying the environment to influence its decisions.  
I3L models assistance via language subgoals by treating the subgoals as extra information added to the environment.
We devise an algorithm for the \iiil setting that yields significant improvements over baselines on the VNLA task on both seen and unseen environments.

The contributions of this paper are: (a) a new task VNLA that represents a step closer to real-world applications of mobile agents accomplishing indoor tasks (b) a novel IL framework that extends the conventional framework to modeling indirect intervention, and (c) a general solution to I3L that is shown to be effective on the VNLA task. The task is accompanied by a large-scale dataset based on the photorealistic Matterport3D environment \cite{chang2017matterport3d,anderson2018vision}.

\section{Related Work}
\textbf{Language and robots.} Learning to translate natural language commands to physical actions is well-studied at the intersection of language and robotics. 
Proposals include a variety of grounded parsing models that are trained from data \cite{matuszek2012joint,matuszek2013learning,krishnamurthy2012weakly} 
and models that interact with robots via natural language queries against a knowledge base \cite{saxena2014robobrain}
Most relevant to the present work are \cite{misra2014tell} who ground natural language to robotic manipulator instructions using Learning-from-Demonstration (LfD) and \cite{duvallet2013imitation} who employ imitation learning of natural language instructions using humans following directions as demonstration data. 
In \cite{hu2019safe}, verbal constraints are used for safe robot navigation in complex real-world environments.  

%\cite{matuszek2013learning} use a semantic parsing model where the grounding relations are learnt from data and is executed on a map of the environment. In \cite{matuszek2012joint} they extend the model to incorporate richer perception models including attribute classifiers. \cite{krishnamurthy2012weakly} utilize a grounded CCG parser to learn a joint model of meaning and context for executing natural language navigational instructions. \cite{misra2014tell} tackle the task of grounding natural language to robotic manipulator instructions using a Learning-from-Demonstration (LfD) approach. \cite{saxena2014robobrain} propose building a rich knowledge-base of multi-modal data like symbols, natural language, visual features etc where the data can come from multiple sources like the internet, interactions with robots and even other research groups. They formulate the task of interacting with robots via natural language as queries to this knowledge base. \cite{duvallet2016inferring} use information embedded in natural language instructions itself as a sensor to infer spatial map layouts and then update their belief over maps as the robot is navigating and sees new information. \cite{duvallet2013imitation} use imitation learning to follow natural language instructions using people following directions as demonstration data. 

%There has been renewed interest in grounded visual navigation from natural language especially given recent advances in reinforcement learning in video and text-based games (\cite{mnih2015human,van2017hybrid,narasimhan2015language}).

\textbf{Simulated environments.} Simple simulators as Puddle World Navigation  \cite{janner2018representation} and Rich 3D Blocks World  \cite{bisk2016natural,bisk2017learning} have facilitated understanding of fundamental representational and grounding issues by allowing for fast experimentation in easily-managed environments. 
Game-based and synthetic environments offer more complex visual contexts and interaction dynamics \cite{kempka2016vizdoom,savva2017minos,brodeur2017home,puig2018virtualhome,dosovitskiy2017carla,wu2018building,babyai_iclr19,cote18textworld}.
Simulators that are more photo-realistic, realistic-physics enabled simulators are beginning to be utilized to train real-world embodied agents \cite{shah2018airsim,chang2017matterport3d,chen2018touchdown,xiazamirhe2018gibsonenv,de2018talk,habitat19arxiv}.

\textbf{End-to-end learning in rich simulators.} \cite{das2018embodied} present the ``Embodied Question Answering'' (EmbodiedQA) task where an agent explores and answers questions about the environment. \cite{das2018neural} propose a hierarchical solution for this task where each level of the hierarchy is independently warmed up with imitation learning and further improved with reinforcement learning. \cite{hermann2017grounded} and \cite{wu2018building} similarly use reinforcement learning in simulated 3D environments for successful execution of written instructions. 
On the vision-language navigation task \cite{anderson2018vision}, cross-modal matching and self-learning significantly improve generalizability to unseen environments \cite{fried2018speaker,wang2018rcm-sil}. 
%\cite{wang2018look} use the dataset of \cite{anderson2018vision} to study the combination of model-based and model-free reinforcement learning methods for navigation from language.

%\cite{} propose a gated-attention mechanism which combines the image and text representations to learn policies using imitation or reinforcement learning in an end-to-end manner.

%\textbf{Imitation learning} \deycomment{Khanh has a nice overview of related work in the ``learning to ask'' tab in the OneNote. Let me (Dey) know if you need a copy of that. But Khanh should take a stab at writing up that paragraph when he has time.}

\textbf{Imitation learning.} Imitation learning \cite{daume2009search,ross2010efficient,ross2011reduction,ross2014reinforcement, chang2015learning,sharaf2017structured} provides an effective learning framework when a teacher for a task is available or can be simulated. 
There has been rich work that focuses on relaxing unrealistic assumptions on the teacher.
\cite{gao2018reinforcement,chang2015learning,nair2018overcoming,hester2017deep,sun2017deeply} study cases where teachers provide imperfect demonstrations.
\cite{zhang2017query,kim2013maximum,judah2014active,laskey2016shiv} construct  policies to minimize the number of queries to the teacher. 
\cite{co2018guiding} provide language instructions at every time step to guide meta-policy learning.
To the best of our knowledge, however, no previous work on imitation learning has explored the case where the agent actively requests changes to the environment to facilitate its learning process.

\section{Vision-based Navigation with Language-based Assistance}
\noindent \textbf{Setup.} Our goal is to train an agent, with vision as the perception modality, that can navigate indoors to find objects by requesting and executing language instructions from humans. 
The agent is able to ``see'' the environment via a monocular camera capturing its first-person view as an RGB image. 
It is also capable of executing language instructions and requesting additional help when in need. The camera image stream and language instructions are the only external input signals provided; the agent is not given a map of the environments or its own location (e.g. via GPS or indoor localization techniques).

The agent starts at a random location in an indoor environment. A \emph{requester} assigns it an object-finding task by sending a high-level \emph{end-goal}, namely to locate an object in a particular room (e.g., ``Find a cup in one of the bathrooms."). The task is always feasible: there is always an object instance in the environment that satisfies the end-goal. The agent is considered to have fulfilled the end-goal if it stops at a location within $d$ meters along the shortest path to an instance of the desired object. Here $d$ is the \emph{success-radius}, a task-specific hyperparameter.
During execution, the agent may get lost and become unable to progress. We enable the agent to automatically sense when this happens and signal an \emph{advisor} for help.\footnote{For simplicity, we assume the advisor has perfect knowledge of the environment, the agent, and the task. In general, as the advisor's main task is to help the agent, perfect knowledge is not necessary. The advisor needs to only possess advantages over the agent (e.g., human-level common sense or reasoning ability, greater experience at indoor navigation, etc.).} 
The advisor then responds with language providing a \emph{subgoal}. The subgoal is a short-term task that is significantly easier to accomplish than the end-goal. 
In this work, we consider subgoals that \emph{describe the next $k$ optimal actions} (e.g. ``Go forward two steps, look right.").
We assume that strictly following the subgoal helps the agent make progress. 

By specifying the agent's task with a high-level end-goal, our setup does \emph{not} assume the requester knows how to accomplish the task before requesting it.
This aspect, along with the agent-advisor interaction, distinguishes our setup from instruction-following setups \cite{anderson2018vision,misra2014tell,misra2018goal,blukis2018mapping,chen2011learning,chen2018touchdown}, in which the requester provides the agent with detailed sequential steps to execute a task only at the beginning.

\noindent \textbf{Constraint formulation.} The agent faces a multi-objective problem: maximizing success rate while minimizing help requests to the advisor. 
Since these objectives are in conflict, as requesting help more often only helps increase success rate, we instead use a hard-constrained formulation: \emph{maximizing success rate without exceeding a budgeted number of help requests}.
%\footnote{An alternative approach is to use a soft-constrained objective: minimizing a weighted sum of success rate and number of help requests. Though equivalent to the hard-constrained formulation (via Lagrange multipliers), the soft-constrained formulation is less interpretable. It is more natural for users to specify $x$ in a hard constraint like ``this agent is not allowed to request help more than $x$ times", than to specify $y$ in a soft constraint like ``this agent is willing to trade off one percent in success rate for $y$ help requests", because the hard-constraint directly quantifies one of the objectives they care about. }
The hard constraint indirectly specifies a trade-off ratio between success rate and help requests.
The problem is reduced to single-objective once the constraint is specified by users based on their preferences.
%The constraint can be flexibly adjusted by users to meet diverse preferences.

\section{Imitation Learning with Indirect Intervention}
\label{sec:i3l}

%Motivated by the VNLA problem, we present Imitation Learning with Indirect Intervention (\iiil), a more general problem.
%I3L is a novel imitation learning setting that models scenarios where a learning agent is monitored by a more qualified expert (e.g., humans) and receives help from him or her through an imperfect communication channel (e.g., through language). 
Motivated by the VNLA problem, we introduce Imitation Learning with Indirect Intervention (\iiil), which models (realistic) scenarios where a learning agent is monitored by a more qualified expert (e.g., a human) and receives help through an imperfect communication channel (e.g., language).

\noindent \textbf{Advisor.} Conventional Imitation Learning (IL) settings \cite{daume2009search,ross2010efficient,ross2011reduction,ross2014reinforcement,chang2015learning,sharaf2017structured,sun2017deeply} involve interaction between a learning agent and a teacher: the agent learns by querying and imitating demonstrations of the teacher. 
In \iiil, in addition to interacting with a teacher, the agent also receives guidance from an \emph{advisor}.
Unlike the teacher, who only interacts with the agent at training time, the advisor assists the agent during both training and test time.

\noindent \textbf{Intervention.} 
The advisor directs the agent to take a sequence of actions through an \emph{intervention}, which can be direct or indirect. 
Interventions are direct when the advisor overwrites the agent's decisions with its own.
By definition, direct interventions are always executed perfectly, i.e. the agent always takes actions the advisor wants it to take. 
%When an intervention is indirect, it can be viewed as actions by an advisor which alter the overall environment of the agent. 
%Once they are introduced, they persist as parts of the environment until they are replaced or removed when the next intervention happens. 
In the case of indirect interventions, the advisor does not ``take over" the agent but instead modifies the environment to influence its decisions.\footnote{The direct/indirect distinction is illustrated more tangibly in a physical agent such as a self-driving car. Turning off automatic driving mode and taking control of the steering wheel constitutes a direct intervention, while issuing a verbal command to stop the car represents an indirect intervention (the command is treated as new information added to the environment).} 
To utilize indirect interventions, the agent must learn to \emph{interpret} them, by mapping them from signals in the environment to sequences of actions in its action space.
This introduces a new type of error into the learning process: intervention interpretation error, which measures how much the interpretations of the interventions diverge from the advisor's original intents. 

\noindent \textbf{Formulation.} 
We assume the environment is a Markov decision process with state transition function $\mathcal{T}$. The agent maintains two policies: a main policy $\pi_{\textrm{main}}$ for making decisions on the main task, and
a help-requesting policy $\pi_{\textrm{help}}$ for deciding when the advisor should intervene. 
We also assume the existence of teacher policies $\pi^*_{\textrm{main}}$ and $\pi^*_{\textrm{help}}$, and an advisor $\Phi$.
Teacher policies are only available during training, while the advisor is always present.
Having a policy $\pi_{\textrm{help}}$ that decides when to ask for help reduces efforts of the advisor to monitor the agent. 
However, it does not prevent the advisor from actively intervening when necessary, because the advisor is able to control $\pi_{\textrm{help}}$'s decisions by modifying the environment appropriately. 
At a state $s_t$, if $\pi_{\textrm{help}}$ decides that the advisor should intervene, the advisor outputs an indirect intervention that directs the agent to take a sequence of actions. In this work, we consider the case when the intervention instructs the agent to take the next $k$ actions $\left( a_t, a_{t+1}, \cdots, a_{t+k-1} \right)$ suggested by the teacher
\begin{align}
    a_{t+i} &= \pi^*_{\textrm{main}}(s_{t+i}), \ \ \ \ \ \ \  \label{eqn:advisor} \\
    s_{t+i+1} &= \mathcal{T}\left(s_{t+i}, a_{t+i}\right) \ \ \ \ 0 \leq i < k \nonumber
\end{align}
The state distribution induced by the agent, $\vec p_{\textrm{agent}}$, depends on both $\pi_{\textrm{main}}$ and $\pi_{\textrm{help}}$.
As in standard imitation learning, in \iiil, the agent's objective is to minimize expected loss on the agent-induced state distribution:
\begin{align}
    \hat{\pi}_{\textrm{main}}, \hat{\pi}_{\textrm{help}} &= \arg\min_{\pi_{\textrm{main}}, \pi_{\textrm{help}}} \mathbb{E}_{s \sim \vec p_{\textrm{agent}}}
    \left[ \mathcal{L} \left( s, \pi_\textrm{main}, \pi_\textrm{help} \right) \right] 
\end{align} where $\mathcal{L}(.,.,.)$ is a loss function. 
%The above objective implies that the two policies can be trained independently using different algorithms. 

\begin{figure}[t]
  \centering
  \includegraphics[width=0.95\linewidth]{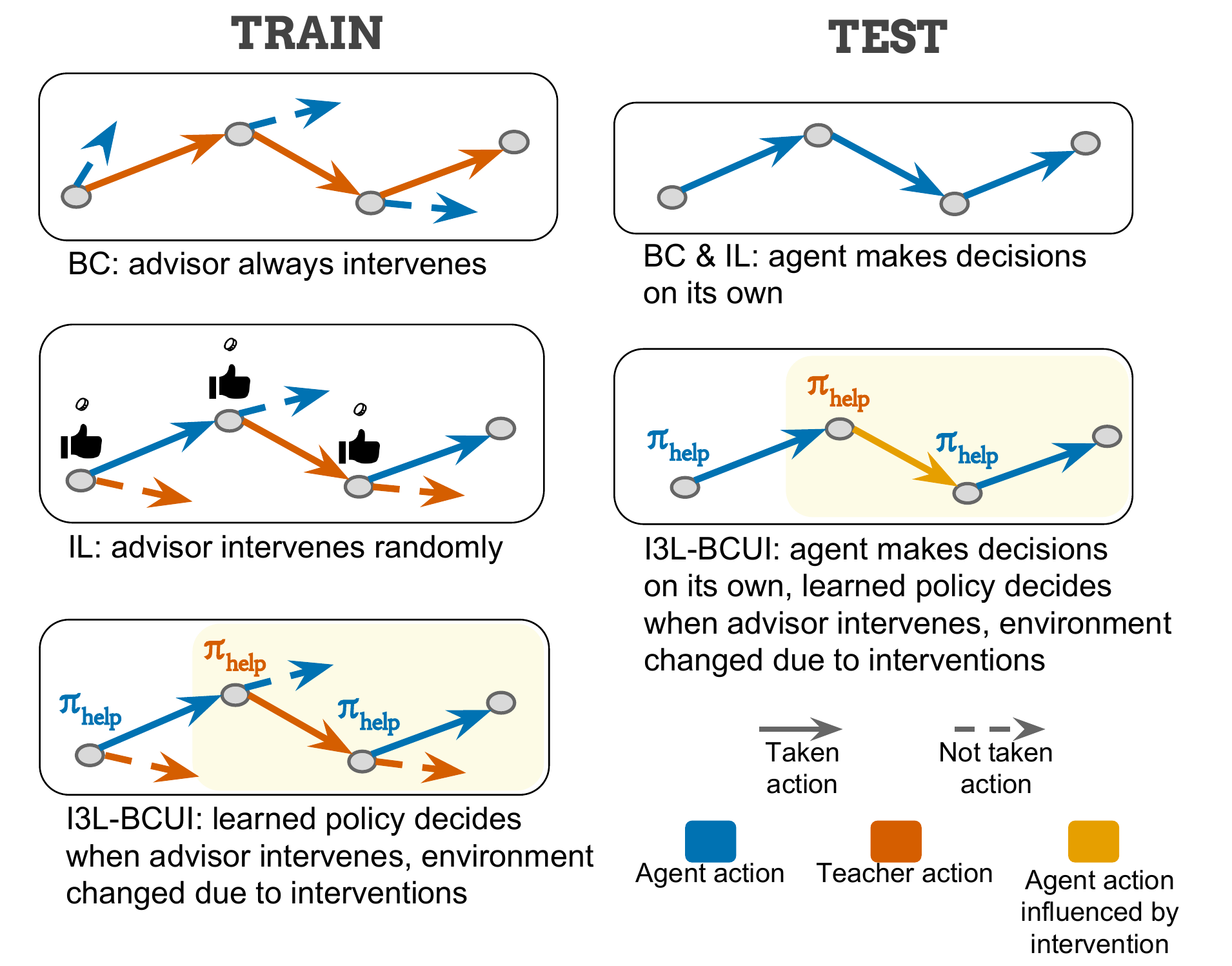}
  \setlength\abovecaptionskip{3pt}
  \caption{Comparison between \iiil trained with behavior cloning under interventions (I3L-BCUI), imitation learning (IL), and behavior cloning (BC) at training time (left) and test time (right). Gray dots represent states and arrows represent actions. Bounding boxes of different colors represent different environments.}
  \label{fig:compare}
\end{figure}

\noindent \textbf{Learning to Interpret Indirect Interventions.}
\iiil can be viewed as an imitation learning problem in a \emph{dynamic} environment, where the environment is altered due to indirect interventions. 
Provided that teacher policies are well-defined in the altered environments, an \iiil problem can be decomposed into a series of IL problems, each of which can be solved with standard IL algorithms. 
It turns out, however, that defining such policies in VNLA is non-trivial.
Even though in VNLA, we use an optimal shortest-path teacher navigation policy, introducing a subgoal to the environment may invalidate this policy.
Suppose when an agent is executing a subgoal, it makes a mistake and deviates from the trajectory suggested by the subgoal (e.g., first turning right for a subgoal ``turn left, go forward"). 
Then, continuing to follow the subgoal is no longer optimal.
Always following the teacher is also not a good choice because the agent may learn to ignore the advisor and not be able to utilize subgoals effectively at test time.

Our solution, which we term as BCUI (\textbf{B}ehavior \textbf{C}loning \textbf{U}nder \textbf{I}nterventions), mixes IL with behavior cloning\footnote{Behavior cloning in IL is equivalent to standard supervised learning in sequence-to-sequence learning, where during training ground-truth tokens (instead of predicted tokens) are always used to transition to the next steps.}.
In this approach, the agent uses the teacher policy as the acting policy (behavior cloning) when executing an intervention ($k$ steps since the intervention is issued).
Thus, the agent never deviates from the trajectory suggested by the intervention and thus never encounters conflicts between the teacher and the advisor.\footnote{A known disadvantage of behavior cloning is that it creates a gap between training and testing conditions, because at test time the agent acts on the learned policy. Addressing this problem is left for future work.}
When no intervention is being executed, the agent uses the learned policy as the acting policy.

\noindent \textbf{Connection to imitation learning and behavior cloning.} Figure \ref{fig:compare} illustrates why I3L trained under BCUI (I3L-BCUI) is a general framework that subsumes both IL and behavior cloning as special cases.
The advisor in I3L-BCUI intervenes both directly (through behavior cloning) and indirectly (by modifying the environment) at training time, 
but intervenes only indirectly at test time. 
The teacher in IL or behavior cloning can be seen as an advisor who is only available during training and intervenes only directly. 
IL and behavior cloning employ simple help-requesting policies. 
In behavior cloning, the help-requesting policy is to always have the teacher intervene, since the agent always lets the teacher make decisions during training. 
Most IL algorithms employ a mixed policy as the acting policy during training, which is equivalent to using a Bernoulli-distribution sampler as the help-requesting policy.
I3L-BCUI imposes no restrictions on the help-requesting policy, which can even be learned from data.

\section{Environment and Data}
\label{sec:env}

\noindent \textbf{Matterport3D simulator.}
The Matterport3D dataset \cite{chang2017matterport3d} is a large RGB-D dataset for scene understanding in indoor environments. It contains 10,800 panoramic views inside 90 real building-scale scenes, constructed from 194,400 RGB-D images. Each scene is a residential building consisting of multiple rooms and floor levels, and is annotated with surface construction, camera poses, and semantic segmentation. Using this dataset, \cite{anderson2018vision} implemented a simulator that emulates an agent walking in indoor environments. The pose of the agent is specified by its viewpoint and orientation (heading angle and elevation angle). Navigation is accomplished by traversing edges in a pre-defined environment graph in which edges connect reachable panoramic viewpoints that are less than 5m apart. 
%We use this simulator to implement VNLI. 
%Models trained in a realistic simulator of this kind are likely to be more amenable transfer to real-world environments than those trained in synthetic simulators \cite{xiazamirhe2018gibsonenv}.

\noindent \textbf{Visual input.}
The agent's pose is not provided as input to the agent. 
Given a pose, the simulator generates an RGB image representing the current first-person view. The image is fed into a ResNet-152 \cite{he2016deep} pretrained on Imagenet \cite{russakovsky2015imagenet} to extract a mean-pooled feature vector, which serves as the input to the agent. We use the precomputed image feature vectors publicly released by \cite{anderson2018vision}.

\noindent \textbf{Action space.} 
%At each viewpoint, the simulator indexes the viewpoint's neighbors in the environment graph starting from one. 
%The set of these indices constitutes the action space of the agent at the current viewpoint. 
%Unfortunately, this action space has dynamic size and each action index does not have a uniform semantic. 
%To alleviate these problems, 
Following \cite{anderson2018vision}, we use a state-independent action space which consists of six actions: \texttt{left}, \texttt{right}, \texttt{up}, \texttt{down}, \texttt{forward} and \texttt{stop}. 
%Details on how we translate an action of one action space to an action of the other are available in the Appendix.
The \texttt{left}, \texttt{right}, \texttt{up}, \texttt{down} actions rotate the camera by $30$ degrees.
The \texttt{forward} action is defined as follows\footnote{Our definition of the \texttt{forward} action, which is different from the one defined in \cite{anderson2018vision}, ensures the navigation teacher never suggests the agent actions that cause it to deviate from the shortest path to the goals.}: executing this action takes the agent to the next viewpoint on the shortest path from the current location to the goals if the viewpoint lies within 30 degrees of the center of the current view, or if it lies horizontally within 30 degrees of the center and the agent cannot bring the viewpoint closer to the center by looking up or down further; otherwise, executing this action takes the agent to the viewpoint closest to the center of the current view. 
We also define a help-requesting action space comprising two actions: \texttt{request} and \texttt{do\_nothing}.

\begin{table}[t]
        \small
        \centering
        \begin{tabular}{lcc}
        \toprule
        \multicolumn{1}{c}{Split} & Number of data points & Number of goals \\ \midrule
        Train & 94,798 & 139,757 \\
        Dev seen & 4,874 & 7,768 \\
        Dev unseen & 5,005 & 8,245\\
        Test seen & 4,917 & 7,470 \\
        Test unseen & 5,001 & 7,537\\ \bottomrule
       \end{tabular}
     \setlength\abovecaptionskip{3pt}  
     \caption{\textsc{AskNav} splits. A data point contains a single starting viewpoint but multiple goal viewpoints.}
     \label{tab:data}
\end{table}

\noindent\textbf{Data Generation.}
Using annotations provided in the Matterport3D dataset, we construct a dataset for the VNLA task, called \asknav. 
We use the same environment splits as \cite{anderson2018vision}: 61 training, 11 development, and 18 test. After filtering out labels that occur less than five times, are difficult to recognize (e.g., ``door frame''), low relevance (e.g., ``window'') or unknown, we obtain 289 object labels and 26 room labels. 
We define each data point as a tuple \emph{(environment, start pose, goal viewpoints, end-goal)}.
%For each environment, we partition all objects into buckets where the instances share the same label and containing room label. 
An end-goal is constructed as ``Find [O] in [R]", where [O] is replaced with ``a/an [object label]" (if singular) or ``[object label]" (if plural), and [R] is replaced with ``the [room label]" (if there is one room of the requested label) or ``one of the \texttt{pluralize}([room label])" (if there are multiple rooms of the requested label). 
%We define the delegate viewpoint of an object as the closest viewpoint that is in the same room. % and is closest to it.
%Goal viewpoints of an end-goal are delegate viewpoints of all object instances in the bucket. 
%All viewpoints in the environment are candidates for the starting viewpoint, except those that are adjacent to, or not reachable from, any goal viewpoint, or that require fewer than 5 or more than 25 actions to reach one of the goal viewpoints.
%Initial orientations are generated as follows: t
%The initial heading angle is a random multiple of $\frac{\pi}{6}$ (less than $2\pi$) and the initial elevation angle is always zero. 
Table \ref{tab:data} summarizes the \asknav dataset.
%We ensure that each pair \emph{(scene ID, end-goal)} appears in exactly one split, and each environment contributes at least one data point (except six environments for which no valid data points exist). 
The development and test sets are further divided into an \emph{unseen} set and a \emph{seen} set. 
The seen set comprises data points that are generated in the training environments but do not appear in the training set.
The unseen set contains data points generated in the development or test environments.
The detailed data generation process is described in the Appendix.

\section{Implementation}
\label{sec:implementation}
\noindent \textbf{Notation.} The agent maintains two policies: a navigation policy $\navpol$ and a help-requesting policy $\askpol$. 
%Both policies takes as input an image of the environment from the current pose $\vec o$, a help-request budget $b$, a language end-goal $\vec g$.
%They are also mutually dependent, as the navigation policy also takes as input a help-requesting action and the help-requesting policy also depends on an input navigation distribution. 
Each policy is stochastic, outputting a distribution $\vec p$ over its action space. 
An action $a$ is chosen by selecting the maximum probability action of or sampling from the output distribution.
The agent is supervised by a navigation teacher $\navpol^*$ and a help-requesting teacher $\askpol^*$ (both are deterministic policies), and is assisted by an advisor $\Phi$.
A dataset $D$ is provided where the $d$-th data point consists of a start viewpoint $\vec x_d^{\textrm{start}}$, a start orientation $\vec \psi_d^{\textrm{start}}$, a set of goal viewpoints $\{ \vec x^{\textrm{end}}_{d, i} \}$, an end-goal $\vec e_{d}$, and the full map $\vec M_d$ of the corresponding environment. 
At any time, the teachers and the advisor have access to the agent's current pose and information provided by the current data point. 

\noindent \textbf{Algorithm.} Algorithm \ref{alg:training} describes the overall procedure for training a VNLA agent. 
We train the navigation policy under the I3L-BCUI algorithm (Section \ref{sec:i3l}) and train the help-requesting policy under behavior cloning. 
%At the start of the training loop for the $d$-th data point, we reset the agent to $(\vec x_d^{\textrm{start}}, \psi_d)$ and initialize the time budget $T$, the help-request constraint $B$, the current help-request budget $q_\textrm{curr}$, and the initial actions $a_{0}^{\textrm{nav}}, a_{0}^{\textrm{ask}}$ (Lines \ref{lst:line:initialize_start}-\ref{lst:line:initialize_end}).  
%initialize the first navigation and help-requesting actions $a_{0}^{\textrm{nav}}, a_{0}^{\textrm{ask*}}$ to the special action $\texttt{<start>}$. 
%We also set the initial end-goal $g_0^{\textrm{main}}$ to the original end-goal $\vec g^{\textrm{main}}_{0,d}$. 
At time step $t$, the agent first receives a view of the environment from the current pose (Line \ref{lst:line:image}). 
It computes a \emph{tentative} navigation distribution $\vec p_{t, 1}^{\textrm{nav}}$ (Line \ref{lst:line:tentative_nav}), which is used as an input to compute a help-requesting distribution $\vec p_t^{\textrm{ask}}$ (Line \ref{lst:line:ask_dist}). 
Since the help-requesting policy is trained under behavior cloning,
the agent invokes the help-requesting teacher $\pi_{\textrm{ask}}^*$ (not the learned policy $\askpol$) to decide if it should request help (Line \ref{lst:line:teacher_ask}). 
%Note that the help-requesting teacher has access to full information of the environment including groundtruth location of the agent $\vec x^{\textrm{curr}}$, groundtruth list of valid end goals $\{ \vec x^{\textrm{end}}_{d, i} \}$ and the obstacle map $\vec M_d$ (Section XXX details our particular implementation of the help-requesting teacher).
If the help-requesting teacher decides that the agent should request help and the help-requesting budget has not been exhausted, the advisor $\Phi$ is invoked to provide help via a language subgoal $\vec g_t^{\textrm{sub}}$ (Lines \ref{lst:line:ask_start}-\ref{lst:line:ask_end}). 
%(Section XXX details our particular advisor implementation.) 
The subgoal is then \emph{prepended} to the original end-goal $\vec g_{0,d}^{\textrm{main}}$ to form a new end-goal $\vec g^{\textrm{main}}_t$ (Line \ref{lst:line:new_goal}). 
If the condition for requesting help is not met, the end-goal is kept unchanged (Line \ref{lst:line:unchanged_goal}). 
After the help-requesting decision has been made, the agent computes a \emph{final} navigation distribution $\vec p_{t, 2}^{\textrm{nav}}$ by invoking the learned policy $\navpol$ the second time.
Note that when computing this distribution, the last help-requesting action is no longer $a^{\textrm{ask}}_{t - 1}$ but has become $a^{\textrm{ask}}_t$.
The agent selects the acting navigation policy based on the principle of the I3L-BCUI algorithm.
Specifically, if the agent has requested help within the last $k$ steps, i.e. it is still executing a subgoal, it uses the teacher policy to act (Line \ref{lst:line:bc_nav}).
Otherwise, it samples an action from the final navigation distribution (Line \ref{lst:line:il_nav}).
In Lines \ref{lst:line:update_start}-\ref{lst:line:update_end}, the learned policies are updated using an online learning algorithm.
Finally, the agent transitions to the next pose according to the taken navigation action (Line \ref{lst:line:transition}). 

%The output of Algorithm \ref{alg:training} is a help-requesting policy $\pi_\textrm{ask}$ and a navigation policy $\pi_\textrm{nav}$, which are used to make decisions at test time. 

%On line 20 the teacher navigation policy $\pi_{\textrm{nav}}^*$ is invoked which similar to $\pi_{\textrm{ask}}^*$ has access to the groundtruth locations and map of the environment. $\pi_{\textrm{nav}}^*$ outputs $a^{\textrm{nav}*}_t$ which is the first step on the shortest path to goal. If help has been requested within the last $k$ steps then the agent is forced to take the navigation teacher's actions directly otherwise an action is sampled from the currently (partially) learnt policy $\pi_{\textrm{nav}}$ which takes as input the current image, language instructions and whether to request for help and outputs a probability distribution over agent action space. In line 26 then the navigation action is executed and the agent transitions to next location. In lines 27, 28 the help-requesting and navigation policies are updated in an online manner. 

\begin{figure}[t]
\centering
  \includegraphics[width=0.95\linewidth]{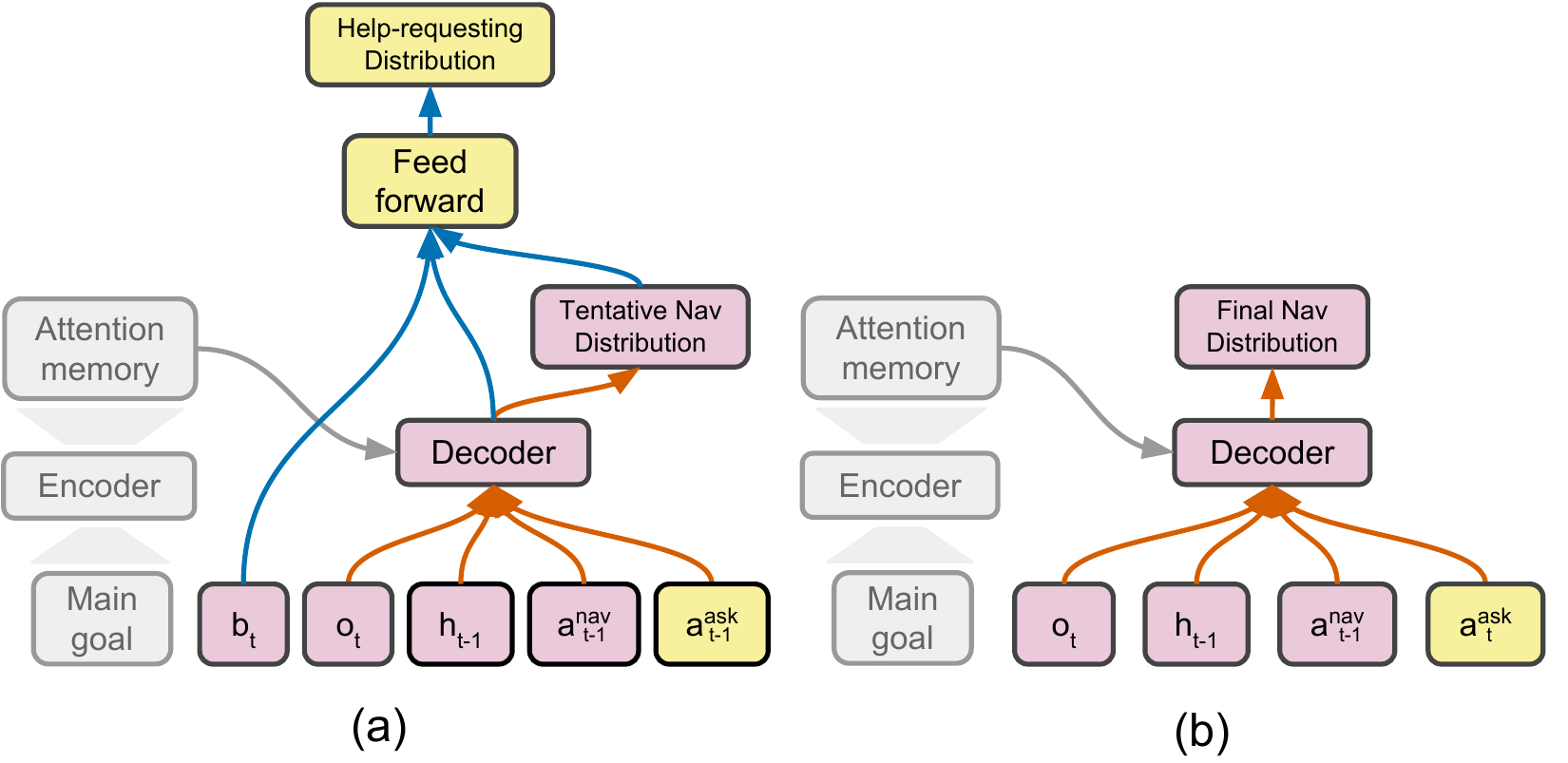}
  \setlength\abovecaptionskip{3pt}
  \caption{Two decoding passes of the navigation module. (a) The first decoding pass computes the tentative navigation distribution, which is used as a feature for computing the help-requesting distribution. (b) The second pass computes the final navigation distribution.}
  \label{fig:two_passes}
\end{figure}

\begin{algorithm}
\caption{VLNA training procedure}
 \label{alg:training}
\begin{algorithmic}[1]
\small
\State Initialize $\pi_{\textrm{nav}}, \pi_{\textrm{ask}}$ randomly.
\State $k$ is the number of next actions a subgoal describes.
%\State Dataset of $D$ examples where each example is defined by a start viewpoint $\vec x^{\textrm{{start}}}_{d}$, start orientation $\vec \psi^{\textrm{start}}_{d}$, goal viewpoints $\{ \vec x^{\textrm{end}}_{d, i} \}$, end-goal $\vec e_d$, and the full map $\vec M_d$ of the corresponding environment.
%\State $k$ is the number of next optimal steps the advisor $\Phi$ describes.
\For{$d = 1 \ldots D$}
\State Reset environment to $(\vec x^{\textrm{{start}}}_{d},  \psi^{\textrm{start}}_{d})$.\label{lst:line:initialize_start}
\State Compute time budget $\hat{T}$ and help-request budget $\hat{B}$. 
\State Initialize current help-request budget $b = \hat{B}$.
\State Initialize $a_{0}^{\textrm{nav}}, a_{0}^{\textrm{ask}}$ to special action $\texttt{<start>}$. \label{lst:line:initialize_end}
\State Initialize $\vec g_0^{\textrm{main}} = \vec e_d$, $\vec x^{\textrm{curr}} = \vec x^{\textrm{{start}}}_{d}$, $\psi^{\textrm{curr}} = \psi^{\textrm{start}}_{d}$. \label{lst:line:initialize_goal}
\For{$t = 1 \ldots \hat{T}$}
\State Receive an image $\vec o_t$ of the current view. \label{lst:line:image}
\State $\vec p^{\textrm{nav}}_{t, 1} = \navpol(\vec o_{t}, \vec g_{t - 1}^{\textrm{main}}, a_{t-1}^{\textrm{nav}}, a_{t-1}^{\textrm{ask}})$ \label{lst:line:tentative_nav}
\State $\vec p^{\textrm{ask}}_t = \askpol(\vec o_t, \vec g_{t - 1}^{\textrm{main}}, \vec p^{\textrm{nav}}_{t, 1}, b)$  \label{lst:line:ask_dist}
\State $a_t^{\textrm{ask}} = a_t^{\textrm{ask}*} = \pi_{\textrm{ask}}^*(\vec p^{\textrm{nav}}_{t, 1}, b, \vec x^{\textrm{curr}}, \psi^{\textrm{curr}}, \{ \vec x^{\textrm{end}}_{d, i} \}, \vec M_d)$ \label{lst:line:teacher_ask}
\If{$b > 0$ and $a_t^{\textrm{ask}} == \texttt{request}$} \label{lst:line:ask_start}
\State $\vec g_t^{\textrm{sub}} = \Phi(\vec x^{\textrm{curr}}, \psi^{\textrm{curr}}, \{ \vec x^{\textrm{end}}_{d, i} \}, \vec M_d, k)$ \label{lst:line:ask_end}
\State $\vec g_t^{\textrm{main}} = \vec g_t^{\textrm{sub}} \odot \vec e_d$ \label{lst:line:new_goal}
\State $b \leftarrow b - 1$
\Else 
\State $\vec g_t^{\textrm{main}} = \vec g_{t-1}^{\textrm{main}}$ \label{lst:line:unchanged_goal}
\EndIf
\State $\vec p^{\textrm{nav}}_{t, 2} = \navpol(\vec o_{t}, \vec g_{t}^{\textrm{main}}, a_{t-1}^{\textrm{nav}}, a_t^{\textrm{ask}})$ \label{lst:line:final_nav}
\State $a^{\textrm{nav}*}_t = \pi_{\textrm{nav}}^*(\vec x^{\textrm{curr}}, \psi^{\textrm{curr}}, \{ \vec x^{\textrm{end}}_{d, i} \}, \vec M_d)$ \label{lst:line:teacher_nav}
\If{$\textrm{requested help within last $k$ steps}$} \label{lst:line:bcui_start}
\State $a^{\textrm{nav}}_t = a^{\textrm{nav}*}_t$ \label{lst:line:bc_nav}
\Else
\State $a_t^{\textrm{nav}} \sim \vec p^{\textrm{nav}}_{t, 2}$ \label{lst:line:il_nav}
\EndIf \label{lst:line:bcui_end}
\State $\askpol \leftarrow \textrm{UpdatePolicy}(\askpol, \vec p^{\textrm{ask}}_t, a_t^{\textrm{ask}*})$ \label{lst:line:update_start}
\State $\navpol \leftarrow \textrm{UpdatePolicy}(\navpol, \vec p^{\textrm{nav}}_{t, 2}, a^{\textrm{nav}*}_t)$ \label{lst:line:update_end}
\If {$a^{\textrm{nav}}_t == \texttt{stop} $}
\State \textbf{break}
\EndIf
\State $\vec x^{\textrm{curr}}, \psi^{\textrm{curr}} \leftarrow \mathcal{T} \left(\vec x^{\textrm{curr}}, \psi^{\textrm{curr}}, a_t^{\textrm{nav}}\right)$ \label{lst:line:transition}
\EndFor
\EndFor 
\end{algorithmic}
\end{algorithm}

\subsection{Agent}

We model the navigation policy $\navpol$ and the help-requesting policy $\askpol$ as two separate neural network modules.
The \textbf{navigation module} is an encoder-decoder model \cite{bahdanau2014neural} with a multiplicative attention mechanism \cite{luong2015effective} and coverage modeling \cite{tu2016modeling}, which encodes an end-goal (a sequence of words) and decodes a sequence of actions. 
Both the encoder and decoder are LSTM-based recurrent neural networks \cite{hochreiter1997long}.
During time step $t$, if the end-goal is updated, the encoder generates an attention memory $\mathcal{M}_t = \left\{ \vec m_1^{\textrm{enc}}, \cdots, \vec m_{|\vec g^{\textrm{main}}_t|}^{\textrm{enc}} \right\} $ by recurrently computing
\begin{align}
	\vec m_i^{\textrm{enc}} &= \textsc{LSTM}_{\textrm{enc}}\left(\vec m_{i - 1}^{\textrm{enc}}, \vec g^{\textrm{main}}_{t, i} \right), \ \ \ 1 \leq i \leq |\vec g^{\textrm{main}}_t| 
\end{align} where $\textsc{LSTM}_{\textrm{enc}}$ is the encoding LSTM, $\vec g^{\textrm{main}}_{t, i}$ is the embedding of the $i$-th word of the end-goal.
Otherwise, $\mathcal{M}_t = \mathcal{M}_{t - 1}$.
The decoder runs \emph{two} forward passes to compute the tentative and the final navigation distributions (Figure \ref{fig:two_passes}). The $i$-th decoding pass proceeds as:
\begin{align}
     \vec h_{t, i}^{\textrm{dec}} &= \textsc{LSTM}_{\textrm{dec}}\left( \vec h_{t - 1, 2}^{\textrm{dec}}, \left[ \vec o_t; \vec a^{\textrm{nav}}_{t - 1}; \bar{\vec a}^{\textrm{ask}}_{t} \right] \right) \\
    \vec h_{t, i}^{\textrm{att}} &= \textsc{Attend}\left( \vec h_{t, i}^{\textrm{dec}}, \mathcal{M}_t \right) \\
    %\vec h^{\textrm{out}}_{t, i} &= \textsc{Tanh} \left( \vec W_o^{\textrm{nav}} \left[ \vec h^{\textrm{att}}_{t, i}; \vec h_{t, i}^{\textrm{dec}} \right] \right) \\
	\vec p^{\textrm{nav}}_{t, i} &= \textsc{Softmax} \left( \vec W_s^{\textrm{nav}} \ \vec h^{\textrm{att}}_{t, i} \right) 
\end{align} where $i \in \{1, 2\}$, $\vec W_s^{\textrm{nav}}$ is learned parameters, $\textsc{Attend}(.,.)$ is the multiplicative attention function, $\vec o_t$ is the visual feature vector of the current view, $\vec a^{\textrm{nav}}_{t - 1}$ is the embedding of the last navigation action, and
\begin{align}
	\bar{\vec a}^{\textrm{ask}}_{t} = \begin{cases}
    	\vec a^{\textrm{ask}}_{t - 1} \ \ \text{ if $i = 1$}, \\
        \vec a^{\textrm{ask}}_{t} \ \ \ \  \text{ if $i = 2$}
    \end{cases}
\end{align} is the embedding of the last help-requesting action. 
%We initialize the decoder with zero vectors instead of the decoder's last hidden states \cite{Britz2017MassiveEO}, i.e. $\vec h_{0, 2}^{\textrm{dec}} = \vec h_{0, 2}^{\textrm{out}} = \vec 0$. 

The \textbf{help-requesting module} is a multi-layer feed-forward neural network with \textsc{ReLU} activation functions and a softmax final layer. Its input features are:
\begin{itemize}[nolistsep]
    \itemsep0em
	\item The visual $\vec o_t$ and the embedding of the current help-request budget $\vec b_t$.
	\item The tentative navigation distribution, $\vec p^{\textrm{nav}}_{t, 1}$.
	\item The tentative navigation decoder states, $\vec h^{\textrm{dec}}_{t, 1}$ and $\vec h^{\textrm{att}}_{t, 1}$.
\end{itemize}  
These features are concatenated and fed into the network to compute the help-requesting distribution 
\begin{align}
    \vec h^{\textrm{ask}}_t &= \textsc{Feed-forward}_l \left( \left[ \vec o_t; \vec b_t; \vec p^{\textrm{nav}}_{t, 1}; \vec h^{\textrm{dec}}_{t, 1};  \vec h^{\textrm{att}}_{t, 1}\right] \right) \\
    \vec p^{\textrm{ask}}_t &= \textsc{Softmax}\left( \vec W_s^{\textrm{ask}} \ \vec h^{\textrm{ask}}_t \right)
\end{align} where $\vec W_s^{\textrm{ask}}$ is a learned parameter and $\textsc{Feed-forward}_l$ is a feed-forward network with $l$ hidden layers. During training, we do not backpropagate errors of the help-requesting module through its input features. Preliminary experiments showed that doing so resulted in lower performance. 
%\deycomment{Why 'preliminary'?}
%\kcomment{Because it was something I remember I did but I don't have the exact numbers now. We can consider it as part of the hyperparameter-tuning process to make a final design choice.}

\subsection{Teachers}
\label{sec:teachers}

\noindent \textbf{Navigation teacher}. 
%At any time, the navigation teacher $\pi_{\textrm{nav}}^*$ has full access to the agent's pose, the goal viewpoints and the environment map. 
The navigation teacher always chooses actions to traverse along the shortest path from the current viewpoint to the goal viewpoints.  
This path is optimal with respect to minimizing the walking distance to the goals, but is not necessarily optimal in the number of navigation actions.
Given an agent's pose, the navigation teacher first adjusts the orientation using the camera-adjusting actions (\texttt{left}, \texttt{right}, \texttt{up}, \texttt{down}) until selecting the \texttt{forward} action advances the agent to the next viewpoint on the shortest path to the goals. 
The teacher issues the \texttt{stop} action when one of the goal viewpoints is reached. 

\noindent \textbf{Help-requesting teacher}. 
%Even with access to full groundtruth information of the agent and the environment, a help-requesting teacher $\pi_{\textrm{ask}}^*$ can be framed in multiple ways and it is unknown a priori at which state an agent should ask for help to result in the optimal usage of the help-requesting budget.
%\kcomment{I don't understand this. Given a fixed nav policy, you CAN compute an optimal help-requesting teacher but it would take too long to do so because you can to basically do dynamic programming over all possible agent's states.}
Even with perfect information about the environment and the agent, computing an optimal help-requesting teacher policy is expensive because this policy depends on (a) the agent's internal state, which lies in a high-dimensional space and (b) the current learned navigation policy, which changes constantly during training. 
We design a heuristic-driven teacher, which decides to request help when:
\begin{enumerate}[label=(\alph*),nolistsep]
\itemsep0em
\item The agent deviates from the shortest path by more than $\delta$ meters. The distance from the agent to a path is defined as the distance from its current viewpoint to the nearest viewpoint on the path. 
\item The agent is ``confused", defined as when the difference between the entropy of the uniform distribution and the entropy of the agent's tentative navigation distribution $\vec p_{t, 1}^{\textrm{nav}}$ is smaller than a threshold $\epsilon$. 
\item The agent has remained at the same viewpoint for the last $\mu$ steps.
\item The help-request budget is greater than or equal to the number of remaining steps. 
%This rule ensures the agent utilizes all help requests in the end.
\item The agent is at a goal viewpoint but the highest-probability action of the tentative navigation distribution is \texttt{forward}. 
\end{enumerate}
Although this heuristic-based teacher may not be optimal, our empirical results show that not only is it effective but it is also easy to imitate. 
Moreover, imitating a clairvoyant teacher is more sample-efficient (theoretically proven \cite{ross2014reinforcement,sun2017deeply}) and results in safer, more robust policies compared to maximizing a reward function with reinforcement learning (empirically shown \cite{choudhury2018data}).
The latter approach imposes weaker constraints on the regularity of the solution and may produce exploitative but unintuitive policies \cite{amodei2016concrete}.

\subsection{Advisor}

Upon receiving a request from the agent, the advisor queries the navigation teacher for $k$ consecutive steps to obtain a sequence of $k$ actions (Equation \ref{eqn:advisor}). 
Next, actions \{\texttt{left}, \texttt{right}, \texttt{up}, \texttt{down}, \texttt{forward}, \texttt{stop}\} are mapped to phrases $\{$``turn left", ``turn right", ``look up", ``look down", ``go forward", ``stop"$\}$, respectively. 
Then, repeated actions are aggregated to make the language more challenging to interpret. 
For example, to describe a turn-right action that is repeated $X$ times, the advisor says ``turn $Y$ degrees right" where $Y = X \times 30$ is the total degrees the agent needs to turn after repeating the turn-right action $X$ times.
Similarly, $Z$ repeated \texttt{forward} actions are phrased as ``go forward $Z$ steps". 
The \texttt{up}, \texttt{down}, \texttt{stop} actions are not aggregated because they are rarely or never repeated. 
Finally, action phrases are joined by commas to form the final subgoal (e.g., ``turn 60 degrees left, go forward 2 steps"). 

\subsection{Help-request Budget}
Let $\hat{T}$ be the time budget and $B$ be the help-request budget. 
Suppose the advisor describes the next $k$ optimal actions in response to each request. We define a hyperparameter $\tau \in [0,1]$, which is the ratio between the total number of steps where the agent receives assistance and the time budget, i.e. $ \tau \equiv \frac{B \cdot k}{\hat{T}}$. Given $\tau$, $\hat{T}$ and $k$, we approximate $B$ by an integral random variable $\hat{B}$
\begin{align}
\label{eqn:q_hat}
	\hat{B} &= \floor{B} + r \\
    r & \sim \textsc{Bernoulli}\left( \{B \} \right) \nonumber \\
    B &= \frac{\hat{T} \cdot \tau}{k} \nonumber
\end{align} where $\{B\} = B - \floor{B}$ is the fractional part of $B$. The random variable $r$ guarantees that $\mathbb{E}_r\left[\frac{\hat{B} \cdot k}{\hat{T}}\right] = \tau$ for a fixed $\hat{T}$ and any positive value of $k$, ensuring fairness when comparing agents interacting with advisors of different $k$s. Due to the randomness introduced by $r$, we evaluate an agent with multiple samples of $\hat{B}$. 
Detail on how we determine $\hat{T}$ for each data point is provided in the appendix.

\section{Experimental Setup}
\noindent \textbf{Baselines.} We compare our learned help-requesting policy (\learntoask) with the following baseline policies:
\begin{itemize}[nolistsep]
    \itemsep0em
	\item \noask: never requests help. 
    \item \askfirst: requests help continuously from the beginning, up to $\hat{B}$. 
    \item \randomask: uniformly randomly chooses $\hat{B}$ steps to request help.
    \item \oracleask: follows the help-requesting teacher ($\askpol^*$). 
\end{itemize}
In each experiment, the same help-requesting policy is used during training and evaluation.

\noindent \textbf{Evaluation metrics}. Our primary metrics are \emph{success rate}, \emph{room-finding success rate}, and \emph{navigation error}. 
Success rate is the fraction of the test set on which the agent successfully fulfills the task. Room-finding success rate is the fraction of the test set on which the agent's final location is in the right room type. 
Navigation error measures the length of the shortest path from the agent's end viewpoint to the goal viewpoints. 
We evaluate each agent with five different random seeds and report means with 95\% confidence intervals.
    
\noindent \textbf{Hyperparameters.} See the Appendix for details.
%The navigation module uses unidirectional single-layer LSTMs as encoder and decoder. 
%The help-requesting module is a feed-forward neural network with one hidden layer. 
%We train the agent with Adam \cite{kingma2014adam} for $10^5$ iterations, using a learning rate of $10^{-4}$ without decaying and a batch size of 100. 
%We regularize the agent with an L2-norm weight of $5\times10^{-4}$ and a dropout ratio of 0.5.
%The help-requesting ratio ($\tau$) is 0.4 and the number of actions suggested by the subgoal advisor ($k$) is 4.
%The deviation threshold ($\delta$), uncertainty threshold ($\epsilon$), and non-moving threshold ($\mu$) are 8, 1.0, and 9, respectively. 
%The success radius ($d$) is fixed at 2 meters. 
%We evaluate each agent with five different random seeds.
%More details about hyperparameters are provided in the Appendix.

\section{Results}
\noindent \textbf{Main results.} Our main results are presented in Table \ref{tab:main}. 
Overall, empowering the agent with the ability to ask for help and assisting it via subgoals greatly boost its performance. 
Requesting help is more useful in unseen environments, improvements over \noask of all other policies being higher on \testunseen than on \testseen.
Even a simple policy like \askfirst yields success rate improvements of 12\% and 14\% over \noask on \testseen and \testunseen respectively. 
The \learntoask policy outperforms all agent-agnostic polices (\noask, \askfirst, \randomask), achieving 9-10\% improvement in success rate over \randomask and 24-28\% over \noask.
An example run of the \learntoask agent is shown in Figure \ref{fig:problem}. 
The insignificant performance gaps between \learntoask and \oracleask indicates that the latter is not only effective but also easy to imitate\footnote{There is a tradeoff between performance and learnability of the help-requesting teacher. By varying hyperparameters, we can obtain a teacher that achieves higher success rate but is harder to imitate.}.
\randomask is largely more successful than \askfirst, hinting that it may be ineffective to request help early too often. 
Nevertheless, \askfirst is better than \randomask at finding rooms on \testseen. 
This may be because on \testseen, although the complete tasks are previously unseen, the room-finding subtasks might have been assigned to the agent during training.
For example, the agent might have never been requested to ``find an armchair in the living room" during training, but it might have been taught to go to the living room to find other objects.
When the agent is asked to find objects in a room it has visited, once the agent recognizes a familiar action history, it can reach the room by memory without much additional assistance.
As the first few actions are crucial, requesting help early is closer to an optimal strategy than randomly requesting in this case. 

\begin{table}[t!]
    \small
    \centering
    \setlength{\tabcolsep}{4pt}
	\begin{tabular}{lccc}
		\toprule
        \multicolumn{1}{c}{} & Success & Room-finding & Mean \\ 
        \multicolumn{1}{c}{\askpol} & rate (\%) $\uparrow$ & success & navigation \\ 
        & & rate (\%) $\uparrow$ & error (m) $\downarrow$ \\ \midrule
        \multicolumn{4}{c}{Test seen} \\ 
        \noask & 28.39 $\pm$ 0.00 & 48.97 $\pm$ 0.00 & 6.29 $\pm$ 0.00 \\
        \askfirst & 40.33 $\pm$ 0.35 & 59.64 $\pm$ 0.22 &  4.36 $\pm$ 0.03 \\
        \randomask & 42.98 $\pm$ 0.44 & 54.61 $\pm$ 0.28 & 4.53 $\pm$ 0.03 \\
        \learntoask & 52.09 $\pm$ 0.13 & 64.84 $\pm$ 0.23 & 3.48 $\pm$ 0.01 \\ \specialrule{.005em}{0.2em}{0.25em}
        \oracleask & \textbf{52.26 $\pm$ 0.16} & \textbf{65.42 $\pm$ 0.25} & \textbf{3.42 $\pm$ 0.01} \\ \midrule
        \multicolumn{4}{c}{Test unseen} \\ 
        \noask & 6.36 $\pm$ 0.00 & 14.34 $\pm$ 0.00 & 11.30 $\pm$ 0.00 \\
        \askfirst & 20.00 $\pm$ 0.10 & 30.23 $\pm$ 0.40 & 7.56 $\pm$ 0.02 \\
        \randomask & 25.05 $\pm$ 0.31 & 33.72 $\pm$ 0.37 & 7.09 $\pm$ 0.05 \\
        \learntoask & 34.50 $\pm$ 0.23 & 44.50 $\pm$ 0.36 & 5.66 $\pm$ 0.02 \\
        \specialrule{.005em}{0.2em}{0.25em} 
        \oracleask & \textbf{34.95 $\pm$ 0.33} & \textbf{44.85 $\pm$ 0.39} & \textbf{5.61 $\pm$ 0.02} \\ \bottomrule
	\end{tabular}
	\smallskip
	\setlength\abovecaptionskip{3pt}
	\caption{Performance of help-requesting policies on \textsc{AskNav} test sets.}
	\label{tab:main}
\end{table}

\begin{table}[t!]
    \small
    \centering
    \setlength{\tabcolsep}{2.5pt}
	\begin{tabular}{lcccc}
		\toprule
        \multicolumn{1}{c}{Advisor} & Subgoals & Train iterations & Test seen & Test unseen \\ \midrule
        Direct & \xmark & 70k & 51.07 $\pm$ 0.17 & 32.19 $\pm$ 0.28 \\
        Direct & \cmark  & 100k & 52.09 $\pm$ 0.13 & 34.56 $\pm$ 0.21 \\ 
        Indirect & \cmark & 100k & 52.09  $\pm$ 0.13 & 34.50 $\pm$ 0.23 \\ \bottomrule
        \end{tabular}
	\smallskip
	\setlength\abovecaptionskip{3pt}
	\caption{Success rates (\%) on \textsc{AskNav} test sets of agents interacting with different advisors. We compare agents that achieve comparable success rates on the \textsc{Dev Seen} split.}
	\setlength\belowcaptionskip{-3pt}  
	\label{tab:subgoal_effect}
\end{table}

\noindent \textbf{Effects of subgoals.} 
Subgoals not only serve to direct the agent, but also act as extra, informative input features. 
We hypothesize that the agent still benefits from receiving subgoals even when it interacts with an advisor who intervenes directly (in which case subgoals seem unneeded). 
To test this hypothesis, we train agents interacting with a \emph{direct} advisor, who overwrites the agents' decisions by its decisions during interventions. 
We consider two variants of this advisor: one responds with a subgoal in response to each help request and the other does not. 
Table \ref{tab:subgoal_effect} compares these with our standard indirect advisor, who at test time sends subgoals but does not overwrite the agent's decisions.
Since success rates on \testseen tend to take a long time to converge, we compare the success rates on \testunseen of agents that have comparable success rates on \textsc{Dev Seen} (the success rates differ by no more than 0.5\%).
The two agents interpreting subgoals face a harder learning problem, and thus require more iterations to attain success rates on \textsc{Dev Seen} comparable to that of the agent not interpreting subgoals.
Receiving subgoals boosts sucess rate by more than 2\% on \testunseen regardless of whether intervention is direct or indirect.

\noindent \textbf{Does the agent learn to identify objects?}
The agent might have only learned to find the requested rooms and have ``luckily" stood close to the target objects because there are only few viewpoints in a room.
To verify if the agent has learned to identify objects after being trained with room type information, we setup a \emph{transfer learning} experiment where an agent trained to fulfill end-goals with room types is evaluated with end-goals \emph{without} room types.
Following a procedure similar the one used to generate the \textsc{AskNav} in Section \ref{sec:env}, we generate a \textsc{NoRoom} dataset, which contains end-goals without room type information. 
Each end-goal in the dataset has the form ``Find [O]", where [O] is an object type. 
Finding any instance of the requested object in any room satisfies the end-goal.
The number of goals in the training split of this dataset is comparable to that of the \textsc{AskNav} dataset (around 140k).
More detail is provided in the Appendix.
The results in Table \ref{tab:no_room} indicate that our agent, equipped with a learned help-requesting policy and trained with room types, learns to recognize objects, as it can find objects without room types significantly better than an agent equipped with a random help-requesting policy and trained specifically to find objects without room types (+10\% on \testseen and +8\% on \testunseen in success rate). 
Unsurprisingly, directly training to find objects without room types yields best results in this setup because training and test input distributions are not mismatched. 

\begin{table}[t!]
    \small
    \centering
    \setlength{\tabcolsep}{4pt}
	\begin{tabular}{lccc}
		\toprule
        \multicolumn{1}{c}{\askpol} & Training set & Test seen & Test unseen \\ \midrule
        \randomask & \textsc{NoRoom} & 43.69 $\pm$ 0.37 & 33.41 $\pm$ 0.39 \\
        \learntoask & \textsc{NoRoom} & 53.71 $\pm$ 0.19 & 44.77 $\pm$ 0.27 \\ 
        \learntoask & \textsc{AskNav} & 53.85 $\pm$ 0.45 & 41.63 $\pm$ 0.24 \\ \bottomrule
        \end{tabular}
	\smallskip
	\setlength\abovecaptionskip{3pt}
	\caption{Success rates (\%) on \textsc{NoRoom} test sets.}
	\label{tab:no_room}
\end{table}

\section{Future Work}
We are exploring ways to provide more natural, fully-linguistic question and answer interactions between advisor and agent, and better theoretical understanding of the \iiil setting and resulting algorithms. We will also be investigating how to transfer from simulators to real-world robots.  

{\small
\bibliographystyle{ieee_fullname}
\bibliography{journal-full,main}
}

\newpage
\setcounter{section}{0}
%%%%%%%%% TITLE
\title{Appendix to \\ Vision-based Navigation with Language-based Assistance \\ via Imitation Learning with Indirect Intervention}

\author{}

\maketitle

\section{Acknowledgements}

We would like thank to Microsoft Research Redmond interns Xin Wang, Ziyu Yao, Shrimai Prabhumoye, Shi Feng, Amr Sharaf, Chen Zhao, and Evan Pu for useful discussions. We are grateful also to Michel Galley and Hal Daum\'{e} III for their advice and support. 
Special thanks goes to Vighnesh Shiv for his meticulous review of the paper. 
We also greatly appreciate the CVPR reviewers for their thorough and insightful comments. 

\section{Data Generation.}

\noindent \textbf{\textsc{AskNav} dataset.} We partition all object instances into buckets where instances in the same bucket share the same environment, containing-room label, and object label. 
For each bucket, an end-goal is constructed as ``Find [O] in [R]", where [O] is replaced with ``a/an [object label]" (if singular) or ``[object label]" (if plural), and [R] is replaced with ``the [room label]" (if there is one room of the requested label) or ``one of the \texttt{pluralize}([room label])"\footnote{We use \url{https://github.com/jazzband/inflect} to check for plurality and perform pluralization.} (if there are multiple rooms of the requested label). 
We define the delegate viewpoint of an object as the closest viewpoint that is in the same room.
To avoid annotation mistakes in the Matterport3D dataset, we ignore object instances that do not lie in the bounding boxes of their rooms.
Goal viewpoints of an end-goal are delegate viewpoints of all object instances in the corresponding bucket. 
All viewpoints in the environment are candidates for the start viewpoint.
We exclude candidates that are not reachable from any goal viewpoint and sample from the remaining candidates at most five start viewpoints per room. The initial heading angle is a random multiple of $\frac{\pi}{6}$ (less than $2\pi$) and the initial elevation angle is always zero. 

Each data point is defined as a tuple \emph{(environment, start pose, goal viewpoints, end-goal)}.
For each data point, we run the navigation teacher to obtain the sequences of actions that take the agent from the start viewpoint to the goal viewpoints.
Note that executing those sequences of actions may not necessarily result in the agent facing the target objects in the end. 
We filter data points whose start viewpoints are adjacent to one of the goal viewpoints on the environment graph, or that require fewer than 5 or more than 25 actions to reach one of the goal viewpoints.
We accumulate valid data points in all seen environments and sample a fraction to construct the seen sets.  
Algorithm \ref{alg:data_sample} describes the sampling procedure.
The remaining data points generated from the training environments form the training set. 
Similarly, the unseen sets are samples of data points in the unseen environments. 
We finally remove examples in the seen sets whose environments do not appear in the training set.

\begin{algorithm}[t!]
\caption{Data sampling procedure.}
 \label{alg:data_sample}
\begin{algorithmic}[1]
\small
\State \textbf{Input}: a set of buckets $B = \{ b_i \}$ where each bucket contains valid data points. $N$ is the maximum number of elements to sample from each bucket ($N=10$ for \textsc{AskNav}, $N=20$ for \textsc{NoRoom}).  
\State \textbf{Output}: a dataset $D$ containing no less than 5000 data points. 
\State Initialize $D = \emptyset$.
\While {$|D| < 5000$ and $B \neq \emptyset$}
    \State Randomly shuffle elements of $B$.
    \State Mark all environments as `not sampled'.
    \For {bucket $b$ in $B$}
        \State $e \leftarrow$ environment of $b$.
        \If {$e$ was `not sampled'}
            \State $s$ is a random sample of at most $N$ elements of $b$.
            \State $D \leftarrow D \cup  s$
            \State Remove $b$ from $B$.
            \State Mark $e$ as `already sampled'.
        \EndIf
    \EndFor
\EndWhile
\end{algorithmic}
\end{algorithm}

\begin{figure*}[t!]
    \centering
    \begin{subfigure}[t]{0.32\textwidth}
        \centering
        \includegraphics[height=\textwidth]{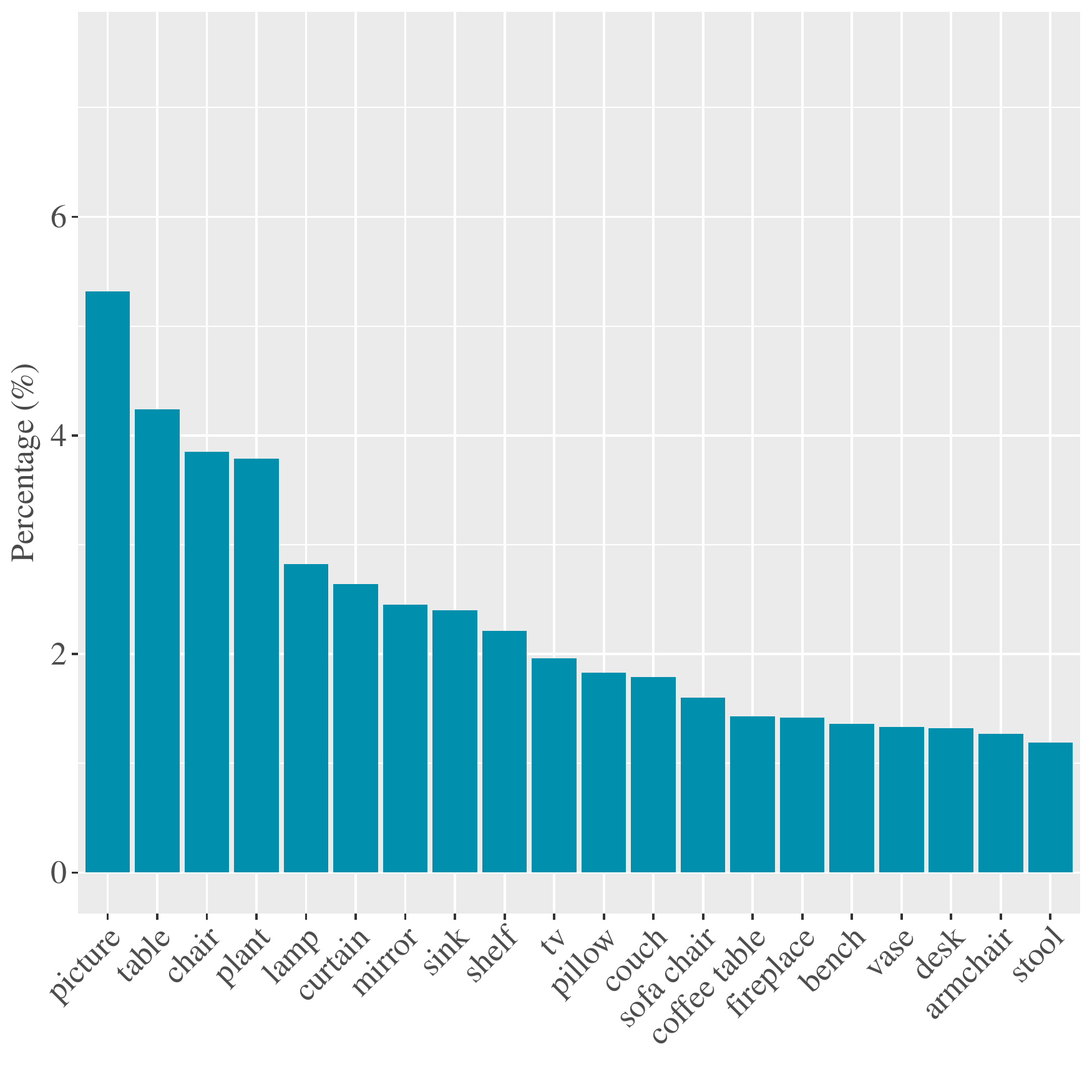}
        \caption{Train}
    \end{subfigure}
    ~ 
    \begin{subfigure}[t]{0.32\textwidth}
        \centering
        \includegraphics[height=\textwidth]{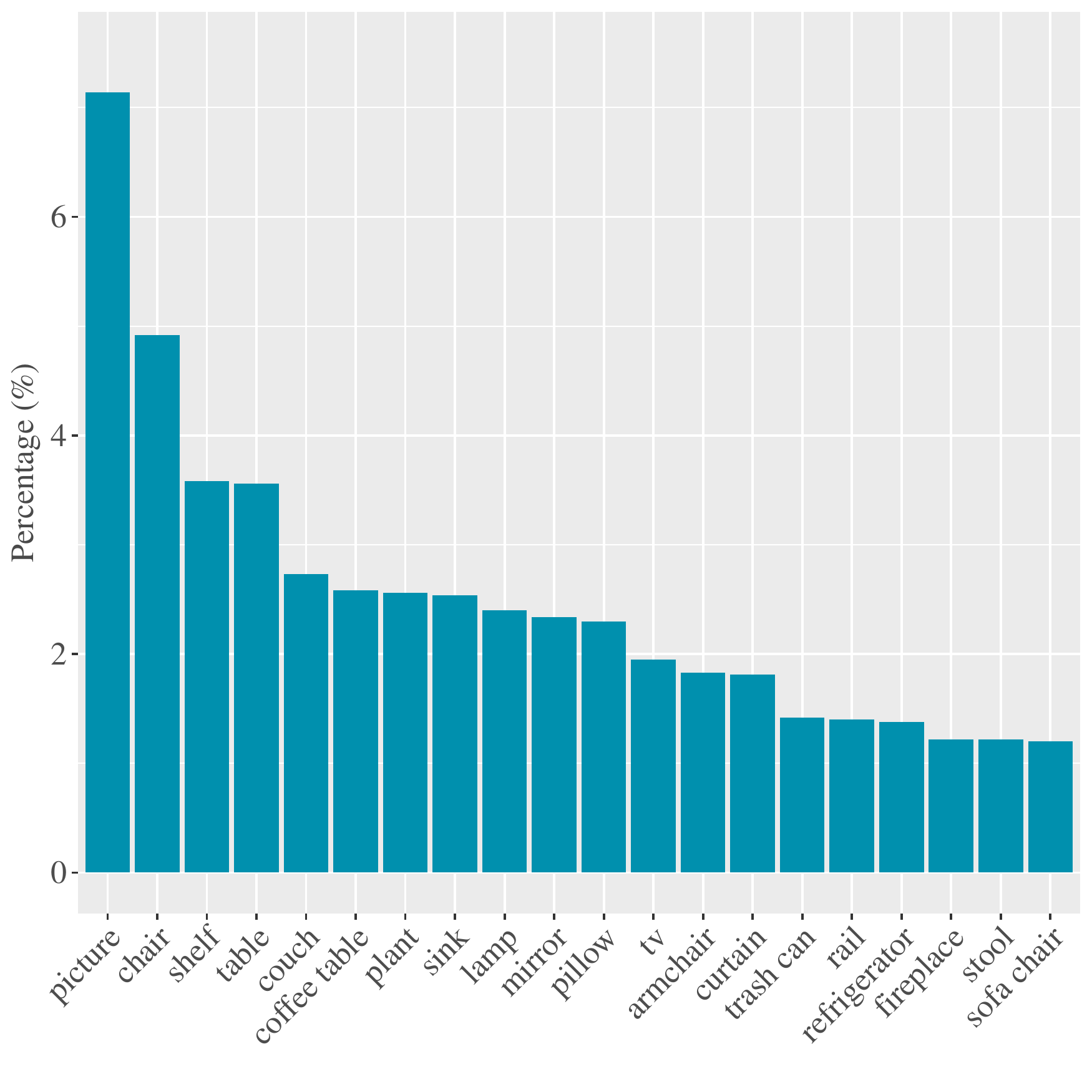}
        \caption{Test seen}
    \end{subfigure}
    ~
    \begin{subfigure}[t]{0.32\textwidth}
        \centering
        \includegraphics[height=\textwidth]{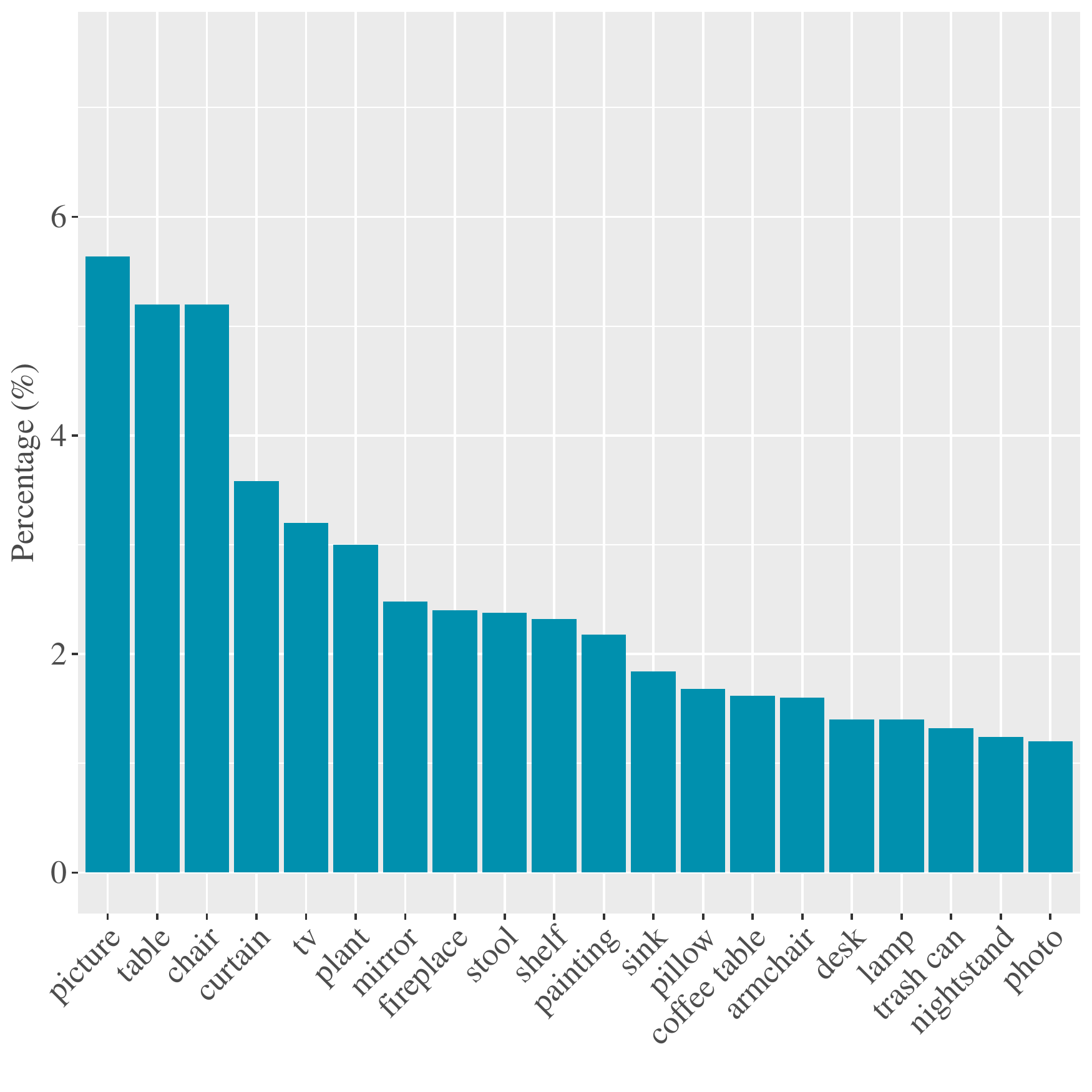}
        \caption{Test unseen}
    \end{subfigure}
    \caption{Top 20 most common objects in the \textsc{AskNav} dataset.}
    \label{fig:obj_dist}
\end{figure*}

\begin{figure*}[t!]
    \centering
    \begin{subfigure}[t]{0.32\textwidth}
        \centering
        \includegraphics[height=\textwidth]{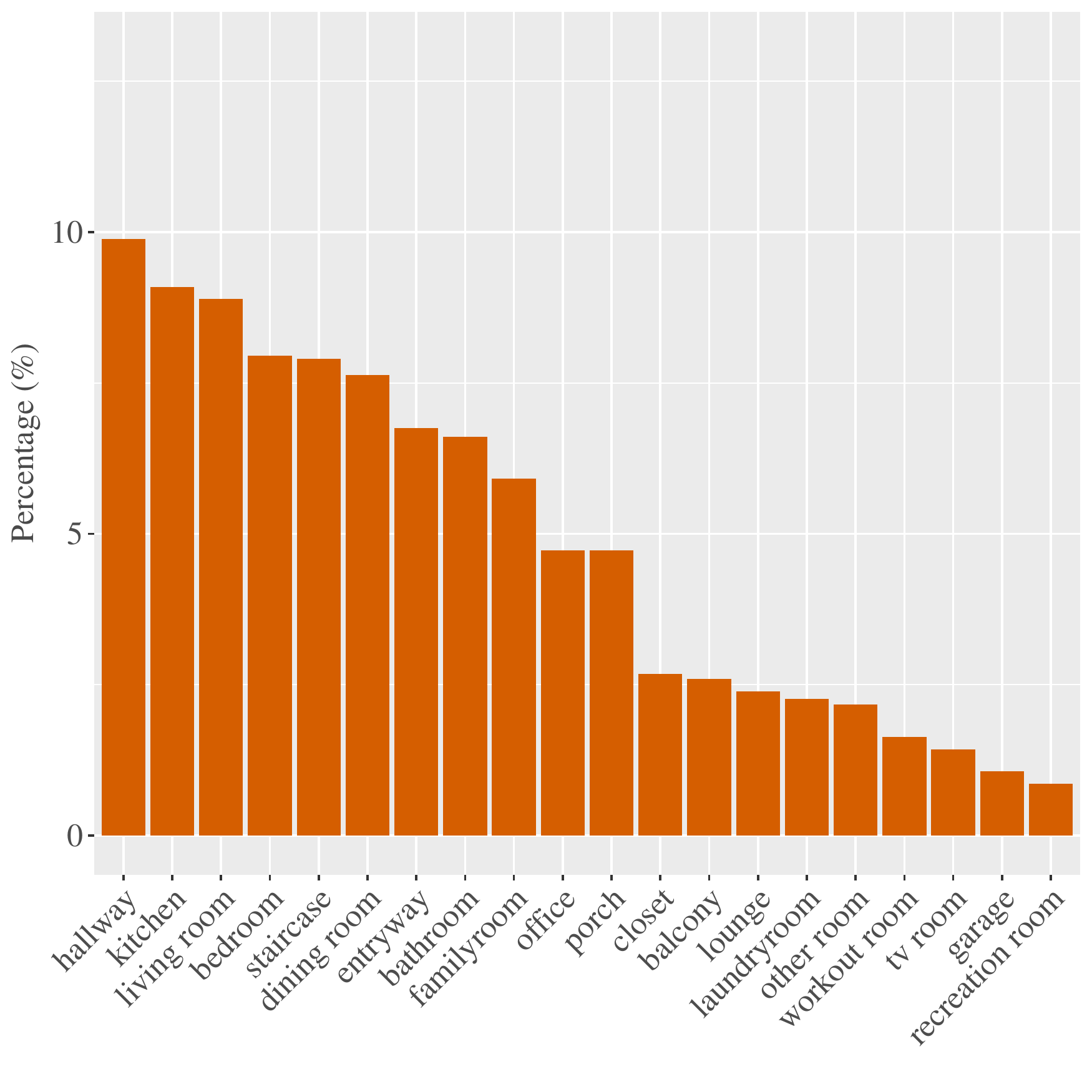}
        \caption{Train}
    \end{subfigure}
    ~ 
    \begin{subfigure}[t]{0.32\textwidth}
        \centering
        \includegraphics[height=\textwidth]{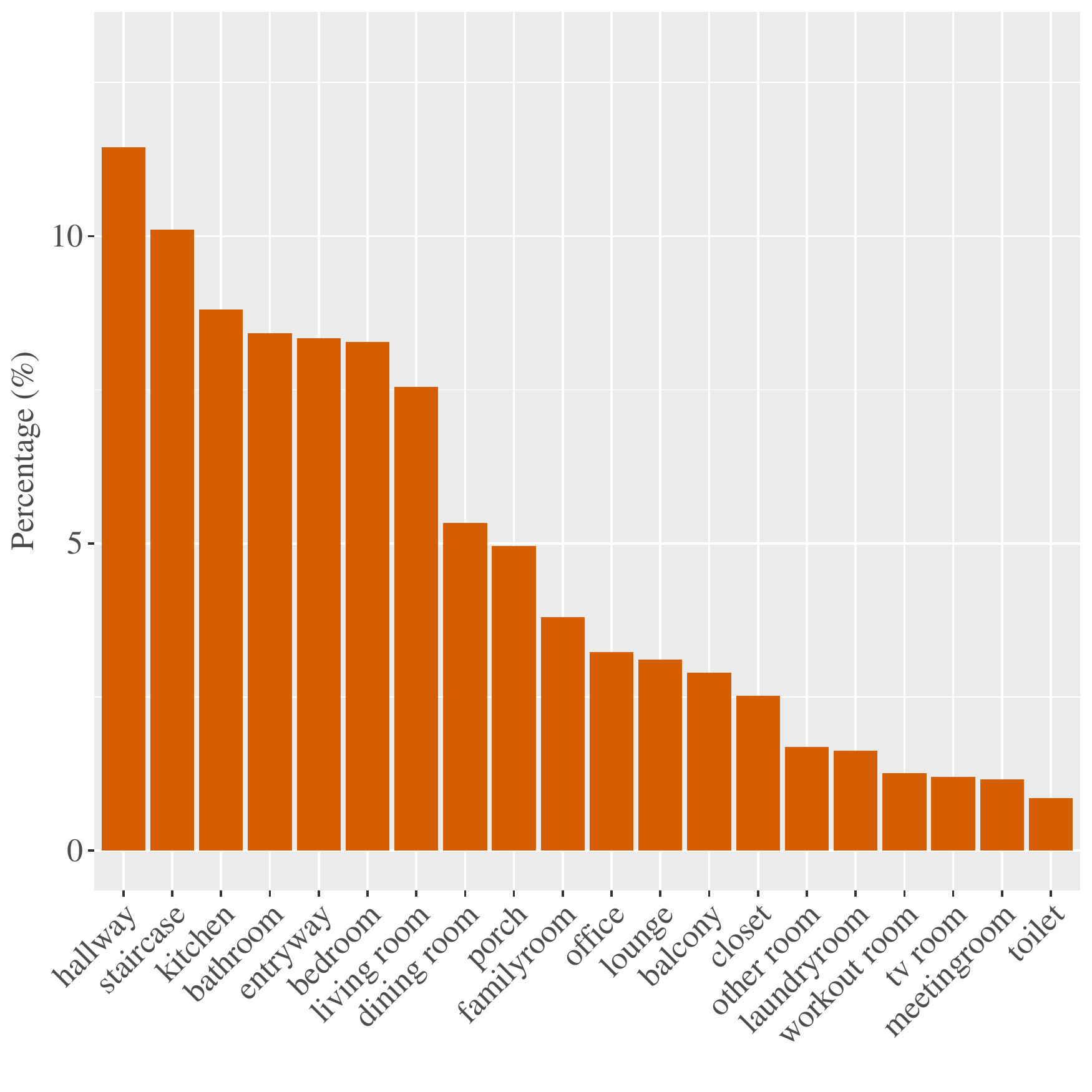}
        \caption{Test seen}
    \end{subfigure}
    ~
    \begin{subfigure}[t]{0.32\textwidth}
        \centering
        \includegraphics[height=\textwidth]{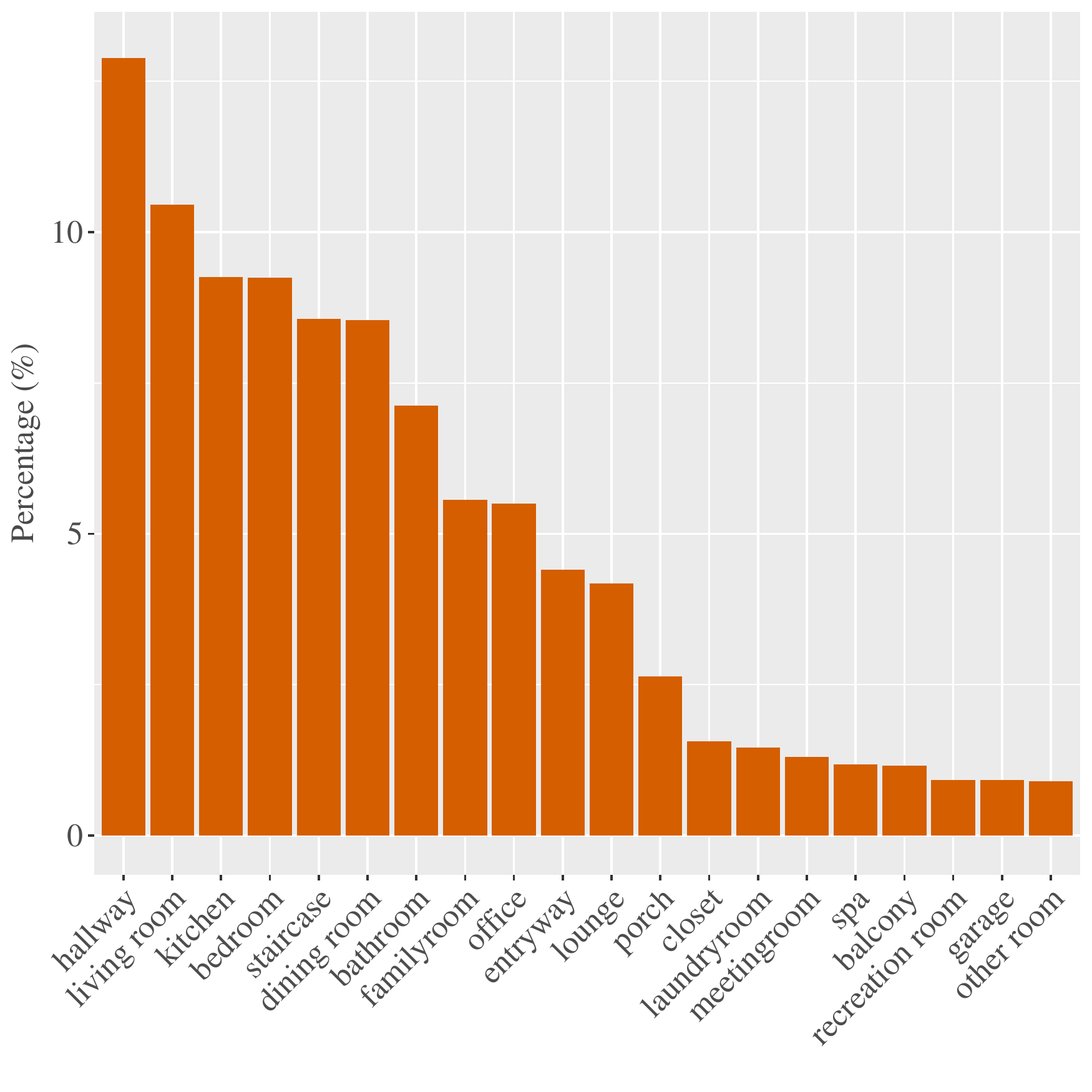}
        \caption{Test unseen}
    \end{subfigure}
    \caption{Top 20 most common start rooms in the \textsc{AskNav} dataset.}
    \label{fig:start_dist}
\end{figure*}

\begin{figure*}[t!]
    \centering
    \begin{subfigure}[t]{0.32\textwidth}
        \centering
        \includegraphics[height=\textwidth]{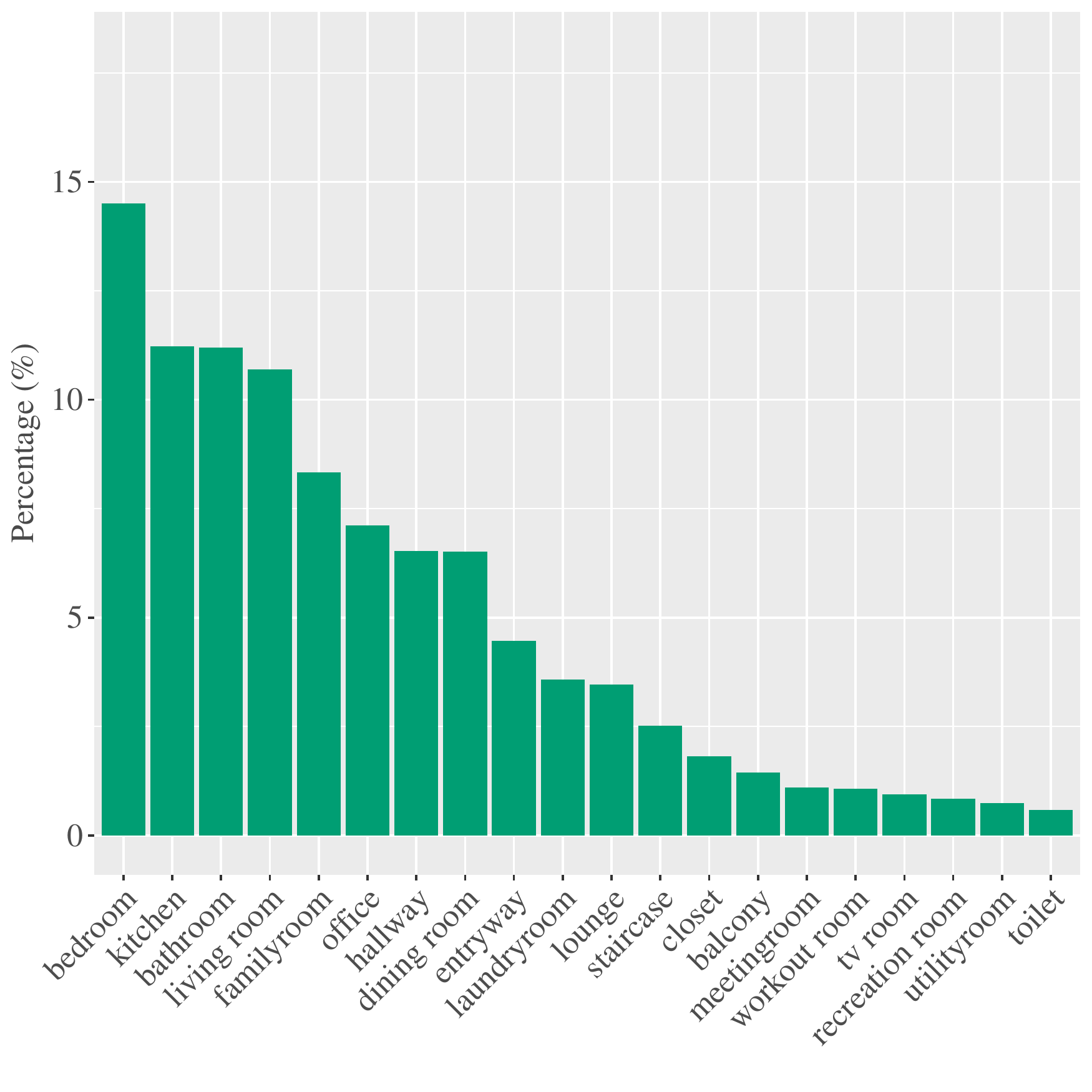}
        \caption{Train}
    \end{subfigure}
    ~ 
    \begin{subfigure}[t]{0.32\textwidth}
        \centering
        \includegraphics[height=\textwidth]{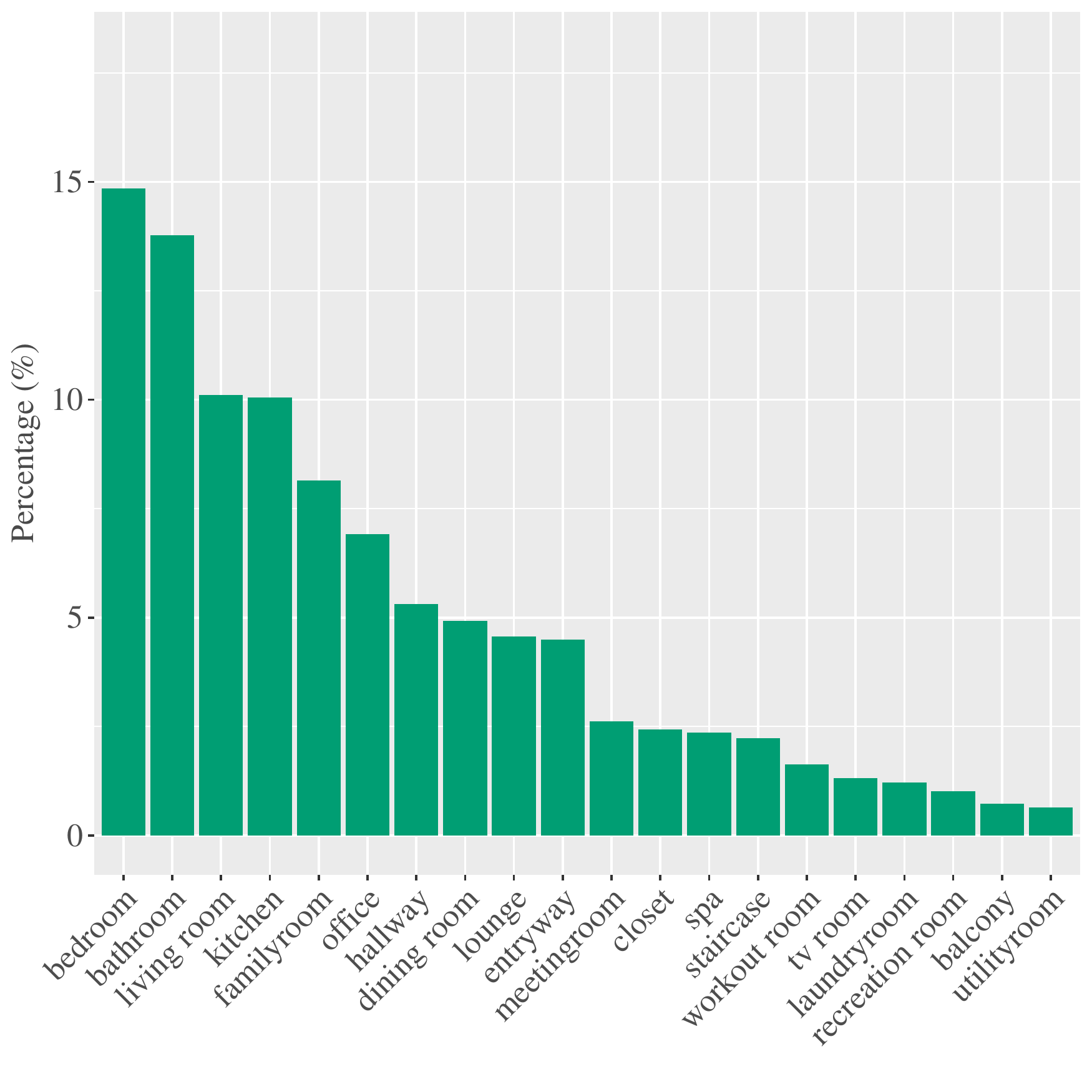}
        \caption{Test seen}
    \end{subfigure}
    ~
    \begin{subfigure}[t]{0.32\textwidth}
        \centering
        \includegraphics[height=\textwidth]{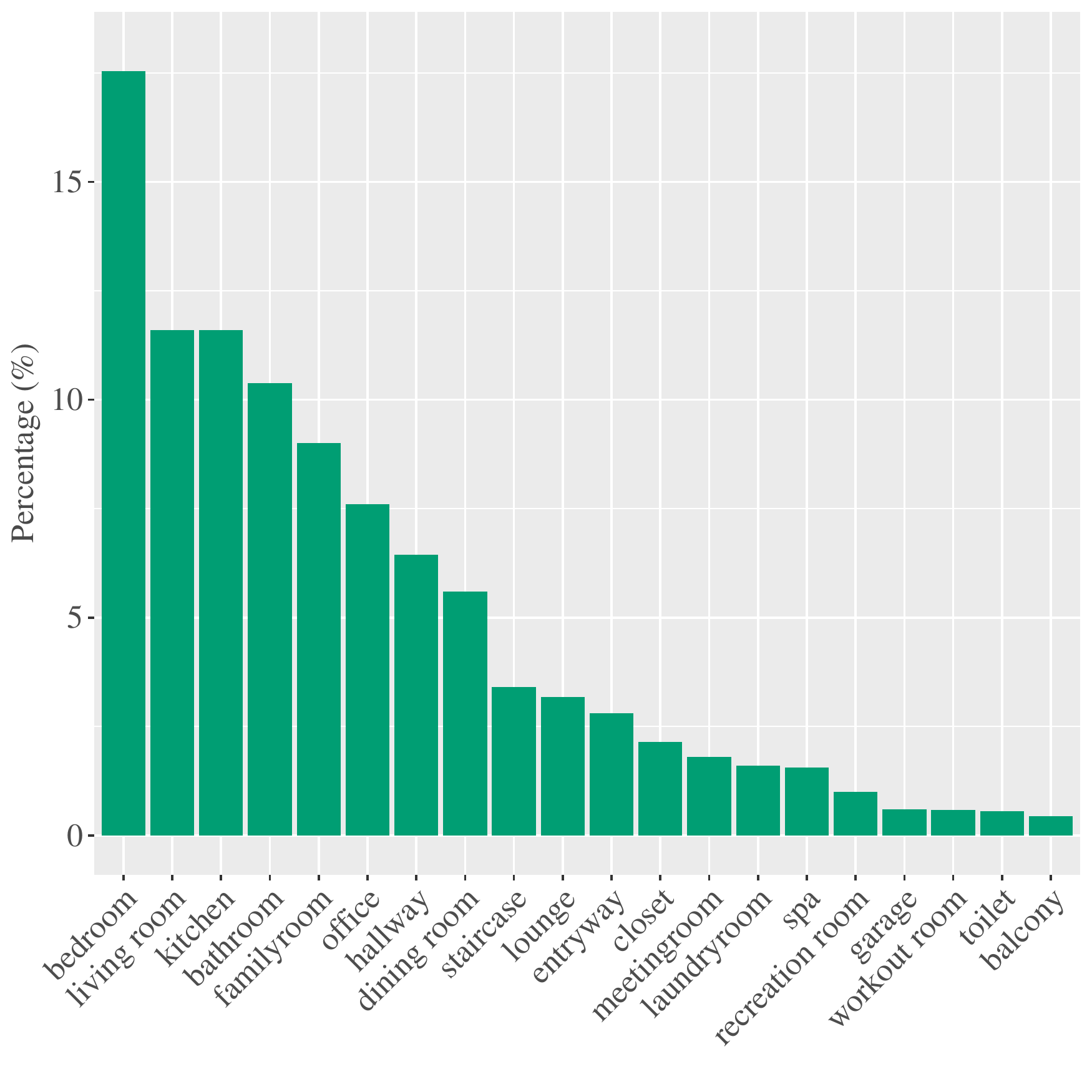}
        \caption{Test unseen}
    \end{subfigure}
    \caption{Top 20 most common goal rooms in the \textsc{AskNav} dataset.}
    \label{fig:goal_dist}
\end{figure*}

\begin{figure}[t!]
    \centering
    \begin{subfigure}[t]{0.3125\linewidth}
        \centering
        \includegraphics[height=\linewidth]{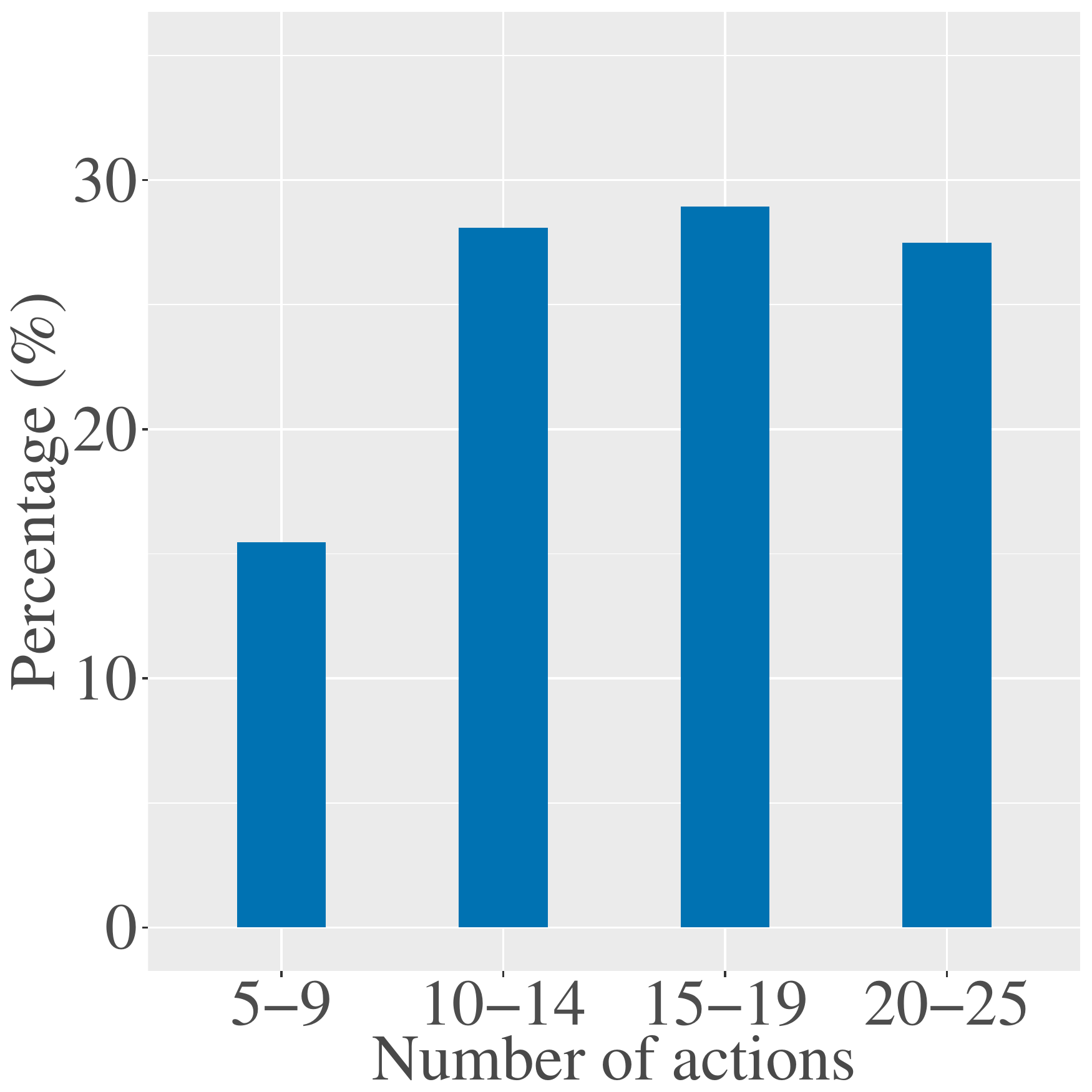}
        \caption{Train}
    \end{subfigure}
    ~ 
    \begin{subfigure}[t]{0.3125\linewidth}
        \centering
        \includegraphics[height=\linewidth]{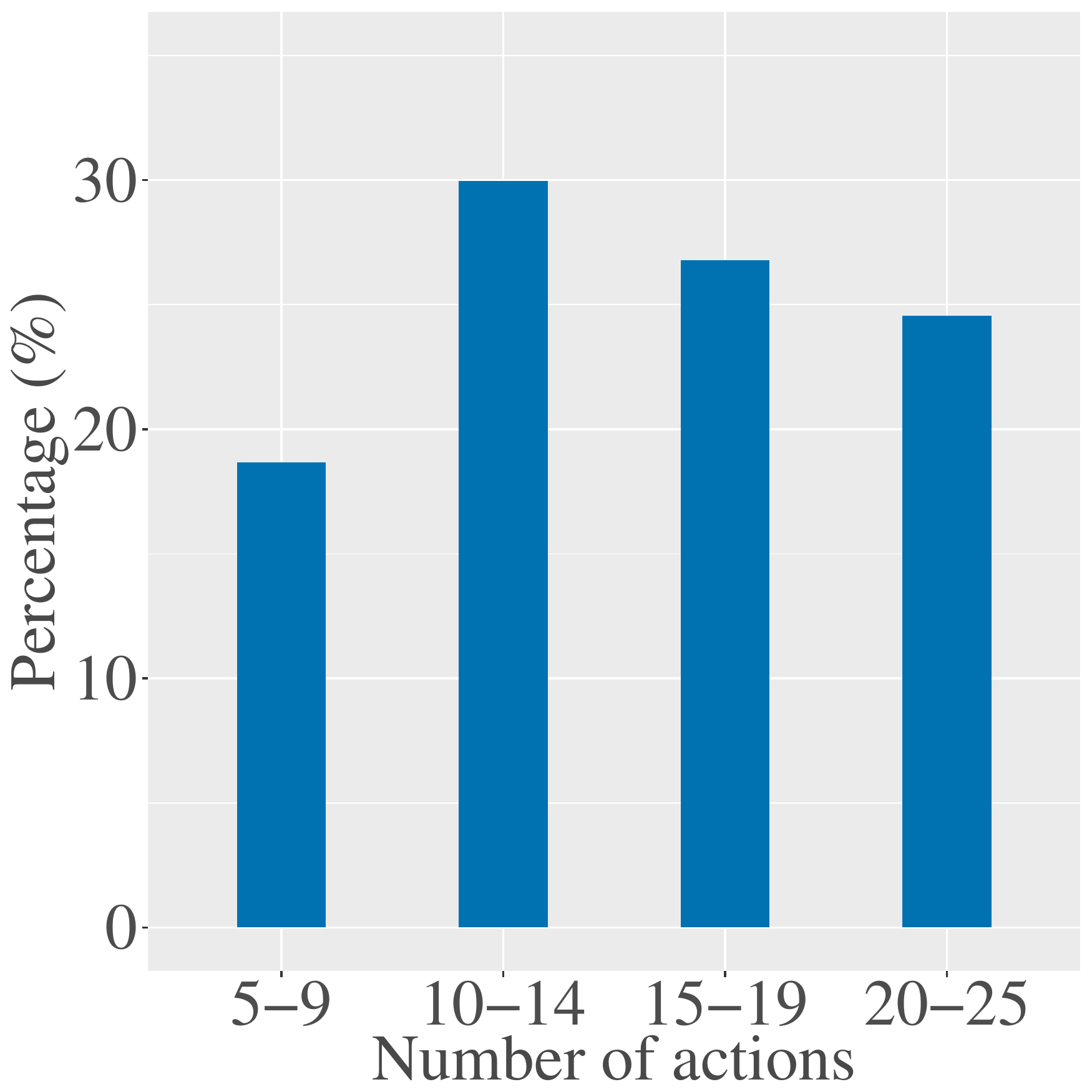}
        \caption{Test seen}
    \end{subfigure}
    ~
    \begin{subfigure}[t]{0.3125\linewidth}
        \centering
        \includegraphics[height=\linewidth]{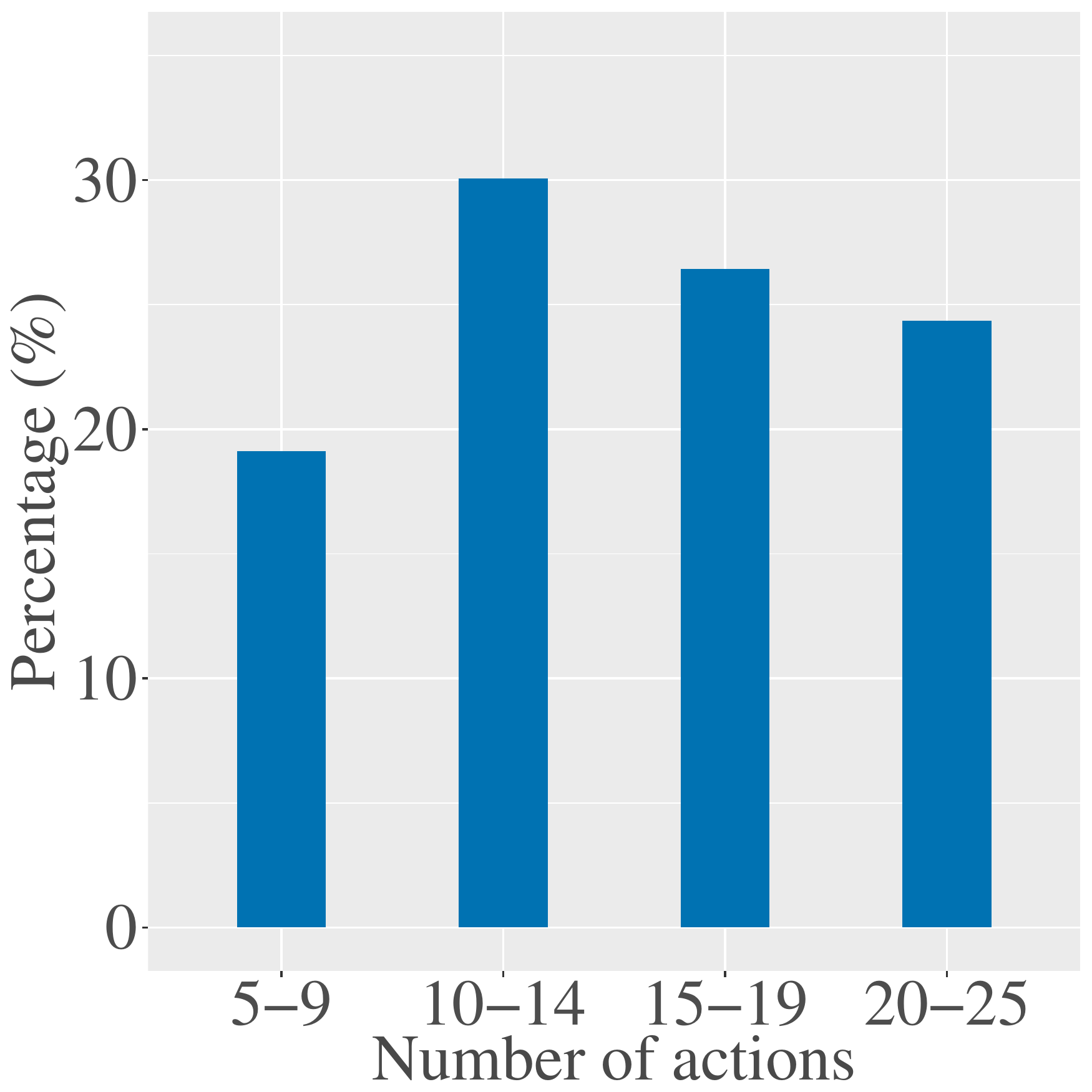}
        \caption{Test unseen}
    \end{subfigure}
    \caption{Distribution of path lengths in the \textsc{AskNav} dataset. Paths are computed by the shortest-path navigation teacher.}
    \label{fig:len_dist}
\end{figure}

Figures \ref{fig:obj_dist}, \ref{fig:goal_dist}, \ref{fig:start_dist}, \ref{fig:len_dist} offer more insights into the \textsc{AskNav} dataset. 
Most common target objects are associated with many instances in a house (e.g., picture, table, chair, curtain).
Goal viewpoints mostly lie in rooms that contain many objects (e.g., bedroom, kitchen, living room, bathroom), whereas start viewpoints tend to be in the hallway because it is spacious and thus includes numerous viewpoints. 
About 85\% of paths require at least ten actions to reach the goal viewpoints. 

\noindent \textbf{\textsc{NoRoom} dataset.} The \textsc{NoRoom} data is generated using a similar procedure described above. 
However, since room types are not provided, each bucket is labeled with only the environment and the object label. 
End-goals have the form ``Find [O]" instead of ``Find [O] in [R]". 
To ensure that the number of goals of this dataset is comparable to that of the \textsc{AskNav} dataset, we sample at most 12 start viewpoints per object for each bucket (we sample five for \textsc{AskNav}).
Table \ref{tab:noroom} summarizes the \textsc{NoRoom} dataset splits. 

\begin{table}[t!]
        \small
        \centering
        \begin{tabular}{lcc}
        \toprule
        \multicolumn{1}{c}{Split} & Number of data points & Number of goals \\ \midrule
        Train & 78,011 & 136,835 \\
        Dev seen & 5,018 & 9,966 \\
        Dev unseen & 5,003 & 9,318 \\
        Test seen & 5,014 & 9,733 \\
        Test unseen & 5,010 & 10,148 \\ \bottomrule
       \end{tabular}
     \caption{\textsc{NoRoom} splits. }
     \label{tab:noroom}
\end{table}

\section{Time budget}
We set the help-request budget $B$ proportional to the time budget $\hat{T}$, which is the (approximate) number of actions required to reach the goal viewpoints. 
During training, for each data point, we set $\hat{T}$ to be the rounded average number of actions needed to move the agent along the shortest path from the start viewpoint to the goal viewpoints.
During evaluation, because the shortest path needs to be unknown to the agent, we compute $\hat{T}$ based on the approximate number of actions to go optimally from the room type of the start viewpoint to the room type of the goal viewpoints\footnote{For the \textsc{NoRoom} dataset, we compute $\hat{T}$ based on the approximate number of actions to go from the start room type to the object type.}. 
This quantity is estimated using the training set. 

Concretely, suppose we evaluate the agent on a data point $d$ with starting viewpoint $\vec x^{\textrm{start}}_d$ and goal viewpoints $\{ \vec x^{\textrm{end}}_{d,i} \}$. 
We define $S$ as the multiset set of numbers of actions of training trajectories whose start and goal room types match those of $d$
\begin{align}
	S = \{ \textsc{TrajLen} \left(\vec x^{\textrm{start}}_{d'}, \{ \vec x^{\textrm{end}}_{d', i'} \} \right) : d' \in D,  \\
    r\left(\vec x^{\textrm{start}}_{d'} \right) = r\left( \vec x^{\textrm{start}}_d \right), 
    r\left(\vec x^{\textrm{end}}_{d', i} \right) = r\left( \vec x^{\textrm{end}}_{d, i} \right) \}
\end{align} where $\textsc{TrajLen}(.,.)$ returns the number of actions to move along the shortest path from a start viewpoint to a set of goal viewpoints, $r(.)$ returns the room type of a viewpoint, and $D$ is the training set. 

Next, we calculate the \emph{95\% upper confidence bound} of the mean number of actions
\begin{align}
\label{eqn:l_hat}
	T &= 
    \begin{cases}
    	\min(c_{\textrm{upper}}, L_{\textrm{max}}), \text{if } \left|S \right| > 0 \\
        L_{\textrm{max}}, \ \ \ \ \ \ \ \ \ \ \ \ \ \ \ \ \ \ \ \ \text{if } \left|S \right| = 0
    \end{cases} \\
    c_{\textrm{upper}} &= \textrm{mean}(S) + 1.95 \cdot \textrm{stdErr}(S) \nonumber
\end{align} $\textrm{mean}(.)$ and $\textrm{stdErr}(.)$ return the mean and standard error of a multiset, respectively, and $L_{\textrm{max}}$ is a pre-defined constant.
We then run the agent for $\hat{T} = \textsc{Round}(T)$ steps.

\begin{table}[t!]
        \small
        \centering
        \begin{tabular}{lc}
        \toprule
        \multicolumn{1}{c}{Hyperparameter} & Value \\ \midrule
        \multicolumn{2}{c}{Navigation module} \\
        LSTM hidden size & 512 \\
        Number of LSTM layers & 1 \\
        Word embedding size & 256 \\
        Navigation action embedding size & 32 \\
        Help-requesting action embedding size & 32 \\ 
        Budget embedding size & 16 \\
        Image embedding size & 2048 \\
        Coverage vector size & 10 \\ \midrule
        \multicolumn{2}{c}{Help-requesting module} \\
        Hidden size & 512 \\
        Number of hidden layers & 1 \\ 
        Activation function & \textsc{ReLU} \\ \midrule
        \multicolumn{2}{c}{Help-requesting teacher} \\
        Deviation threshold ($\delta$) & 8 \\
        Uncertainty threshold ($\epsilon$) & 1.0 \\
        Non-moving threshold ($\mu$) & 9 \\
        Number of actions suggested by a subgoal ($k$) & 4 \\ \midrule
        \multicolumn{2}{c}{Training} \\
        Optimizer & Adam \\
        Number of training iterations & $10^5$ \\
        Learning rate & $10^{-4}$ \\ 
        Learning rate decay & No \\
        Batch size & 100 \\
        Weight decay (L2-norm regularization) & $5 \times 10^{-4}$ \\
        Dropout ratio & 0.5 \\
        Help-requesting ratio ($\tau$) & 0.4 \\ \midrule
        \multicolumn{2}{c}{Evaluation} \\
        Success radius (d) & 2 \\
        Number of evaluating random seeds & 5 \\
        Maximum time budget ($L_{\textrm{max}}$) & 25 \\
        \bottomrule
       \end{tabular}
     \caption{Hyperparameters. }
     \label{tab:hyper}
\end{table}

\section{Hyperparameters}

Table \ref{tab:hyper} summarizes hyperparameters used in our experiments. 
The navigation module uses unidirectional single-layer LSTMs as encoder and decoder. 
We initialize the encoder and the decoder by zero vectors.
The help-requesting module is a feed-forward neural network with one hidden layer. 
We train the agent with Adam \cite{kingma2014adam} for $10^5$ iterations, using a learning rate of $10^{-4}$ without decaying and a batch size of 100. 
We regularize the agent with an L2-norm weight of $5\times10^{-4}$ and a dropout ratio of 0.5.
Training a \textsc{Learned} model took about 17 hours on a machine with a Titan Xp GPU and an Intel 4.00GHz CPU.
The help-requesting ratio ($\tau$) is 0.4 and the number of actions suggested by the subgoal advisor ($k$) is 4.
The deviation threshold ($\delta$), uncertainty threshold ($\epsilon$), and non-moving threshold ($\mu$) are 8, 1.0, and 9, respectively. 
The success radius ($d$) is fixed at 2 meters. 
Both the navigation and help-requesting modules are trained under the maximum log-likelihood objective, which maximizes the model-estimated probabilities of actions suggested by the teacher.
We evaluate each agent with five different random seeds.

\section{Qualitative Analysis}

\begin{figure}[t]
        \centering
        \includegraphics[height=0.9\linewidth]{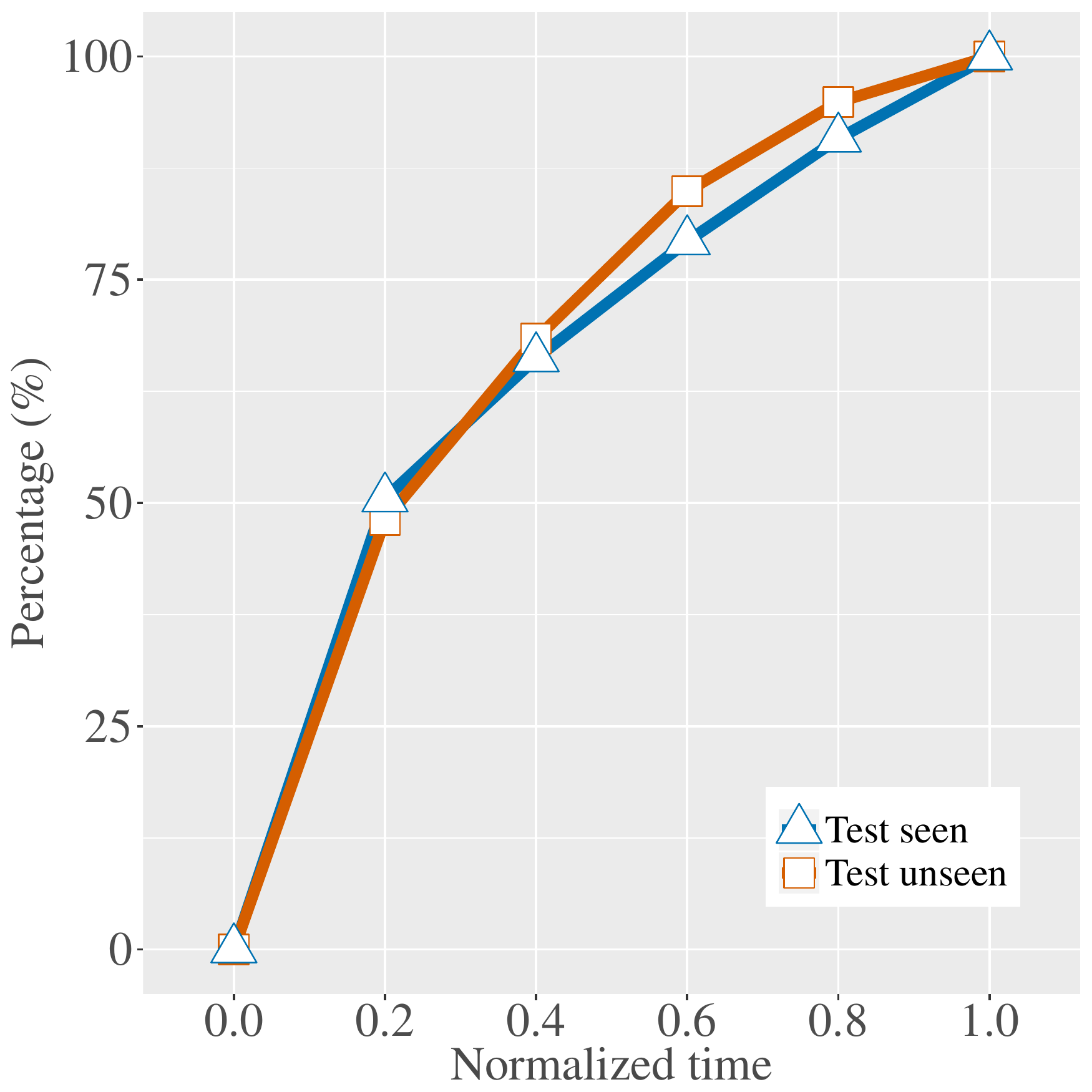}
        \caption{Fraction of help requests made over (normalized) time. }
        \label{fig:request_dist}
\end{figure}

\begin{figure}[t!]
    \centering
    \begin{subfigure}[t]{0.48\linewidth}
        \centering
        \includegraphics[height=\linewidth]{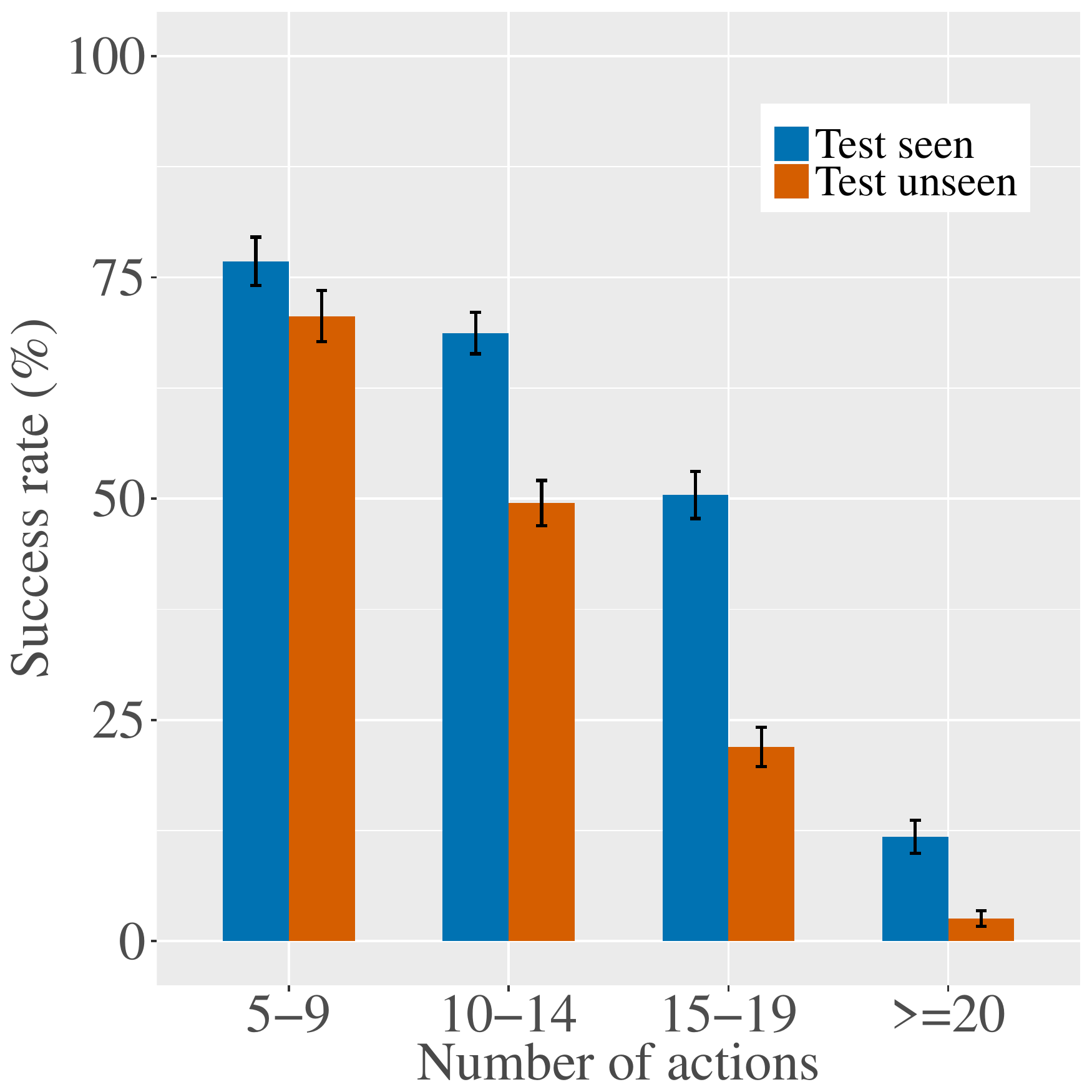}
        \caption{}
        \label{fig:success_optimal}
    \end{subfigure}
    ~
    \begin{subfigure}[t]{0.48\linewidth}
        \centering
        \includegraphics[height=\linewidth]{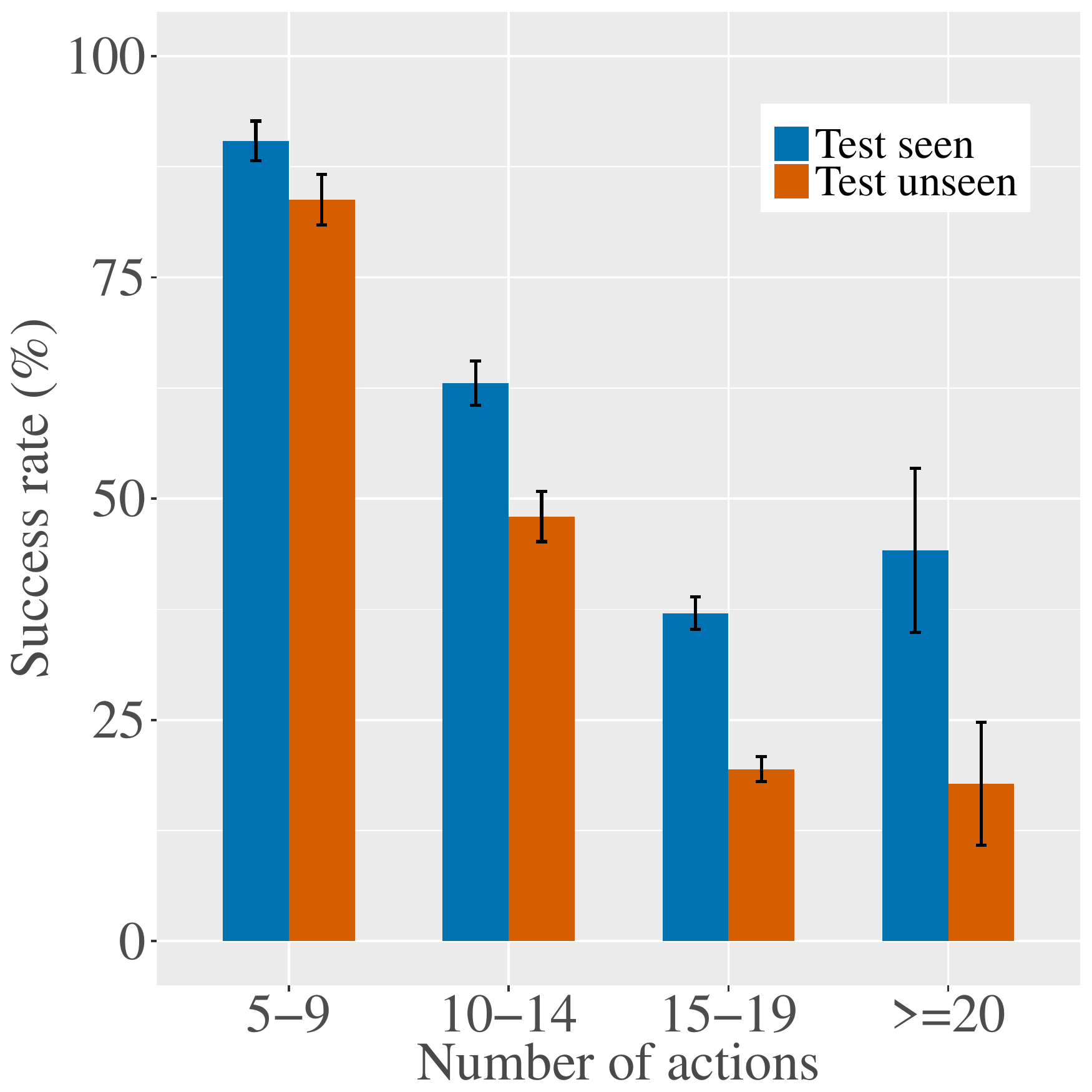}
        \caption{}
        \label{fig:success_predict}
    \end{subfigure}
    \caption{Success rate versus number of actions taken by (a) the navigation teacher and (b) the agent. Error bars are 95\% confidence intervals.}
    \label{fig:success_len}
\end{figure}

\begin{figure}[t!]
    \centering
    \begin{subfigure}[t]{0.48\linewidth}
        \centering
        \includegraphics[height=\linewidth]{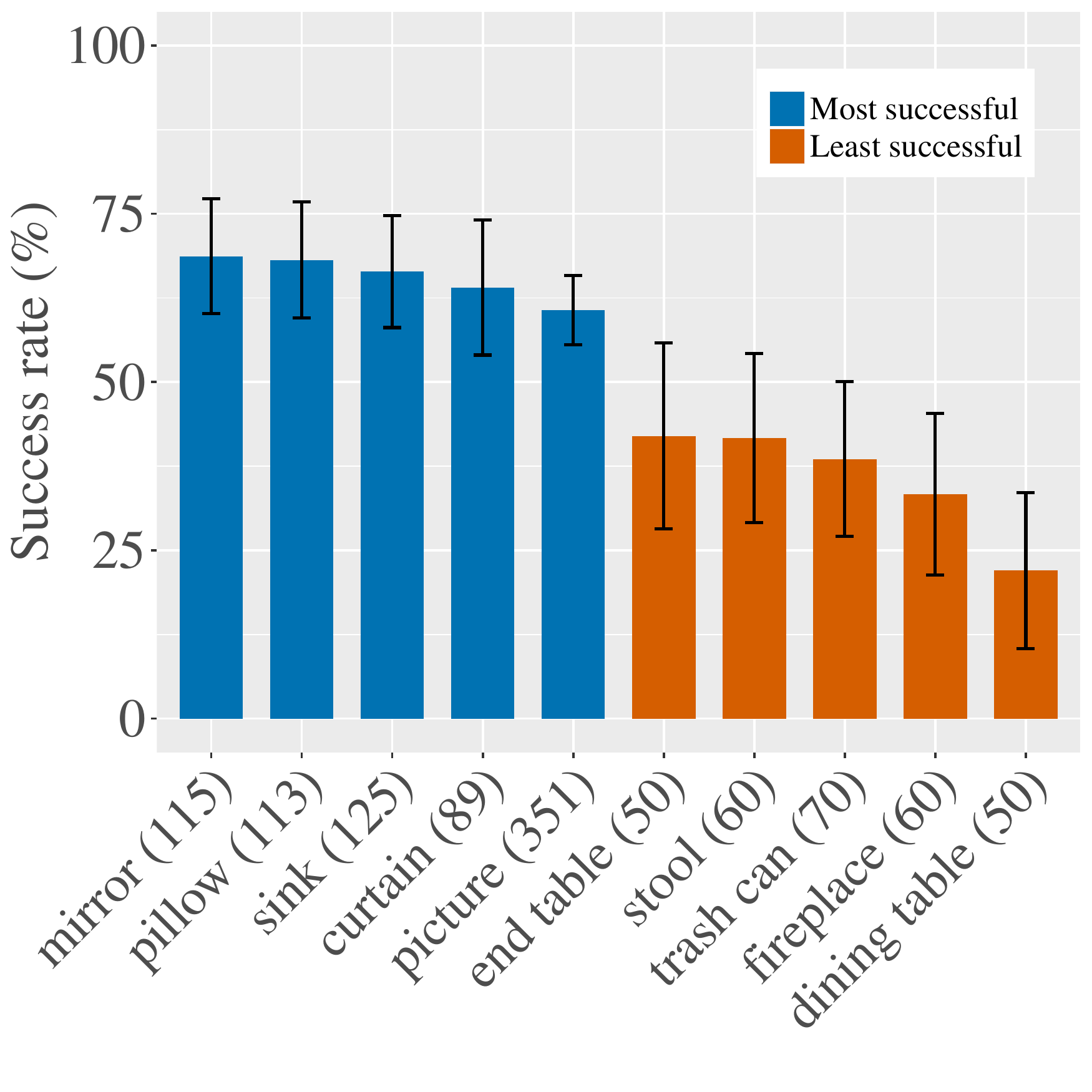}
        \caption{}
        \label{fig:success_obj_seen}
    \end{subfigure}
    ~
    \begin{subfigure}[t]{0.48\linewidth}
        \centering
        \includegraphics[height=\linewidth]{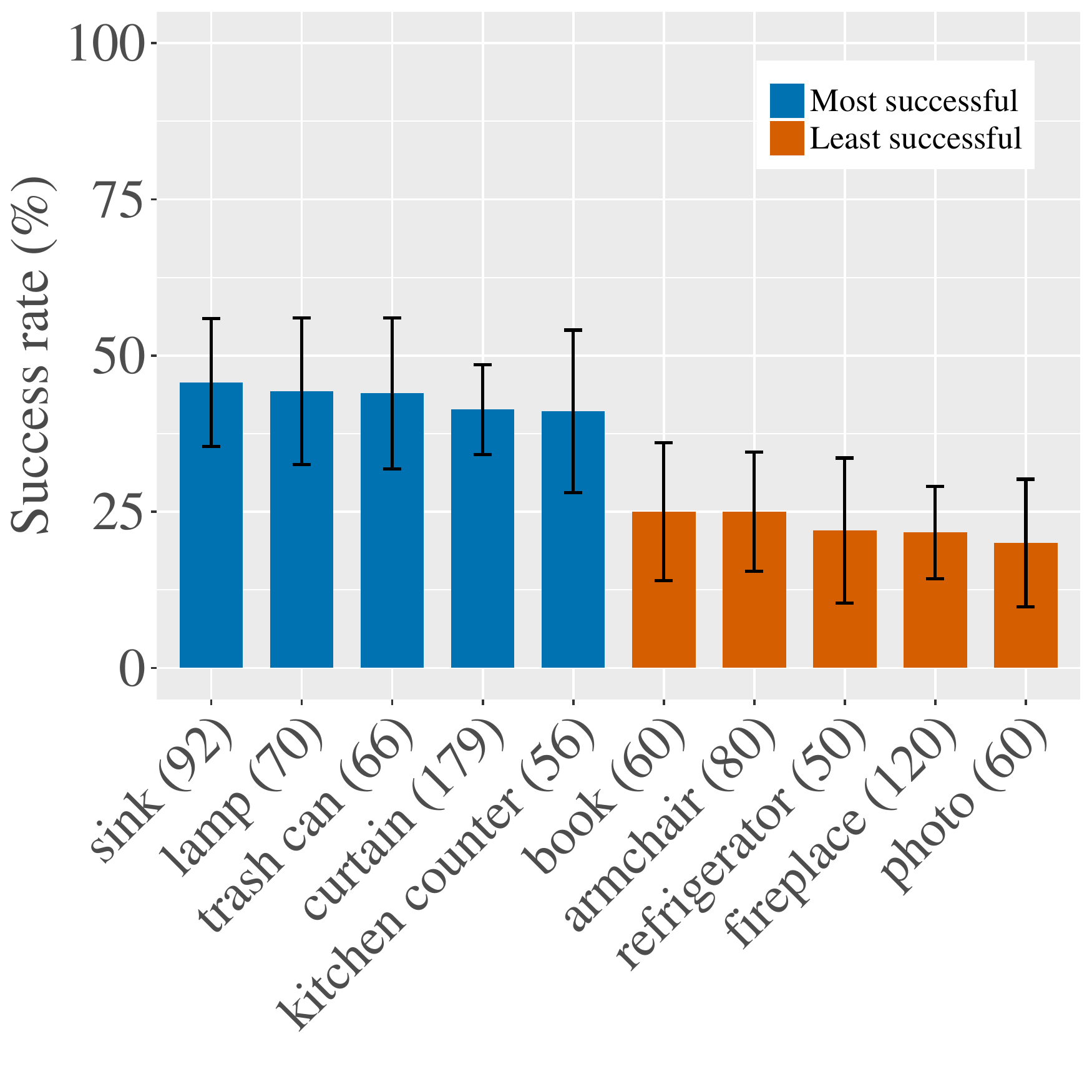}
        \caption{}
        \label{fig:success_obj_unseen}
    \end{subfigure}
    \caption{Top five objects with highest and lowest average success rates in (a) \textsc{Test seen} and (b) \textsc{Test unseen}. Numbers in parentheses are object frequencies. Only objects appearing more than 50 times are included. Error bars are 95\% confidence intervals.}
    \label{fig:success_obj}
\end{figure}

\begin{figure}[t!]
    \centering
    \begin{subfigure}[t]{0.48\linewidth}
        \centering
        \includegraphics[height=\linewidth]{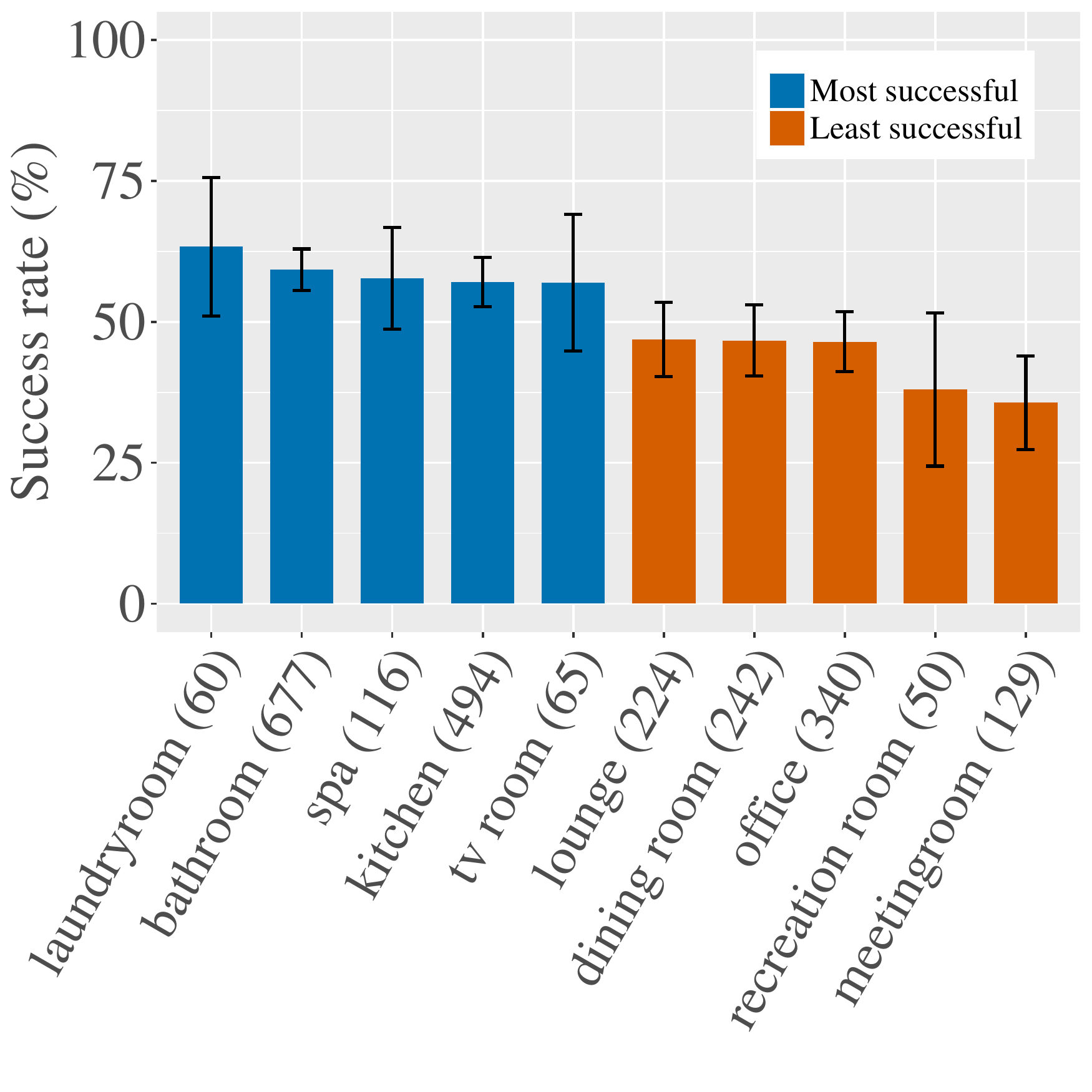}
        \caption{}
        \label{fig:success_room_seen}
    \end{subfigure}
    ~
    \begin{subfigure}[t]{0.48\linewidth}
        \centering
        \includegraphics[height=\linewidth]{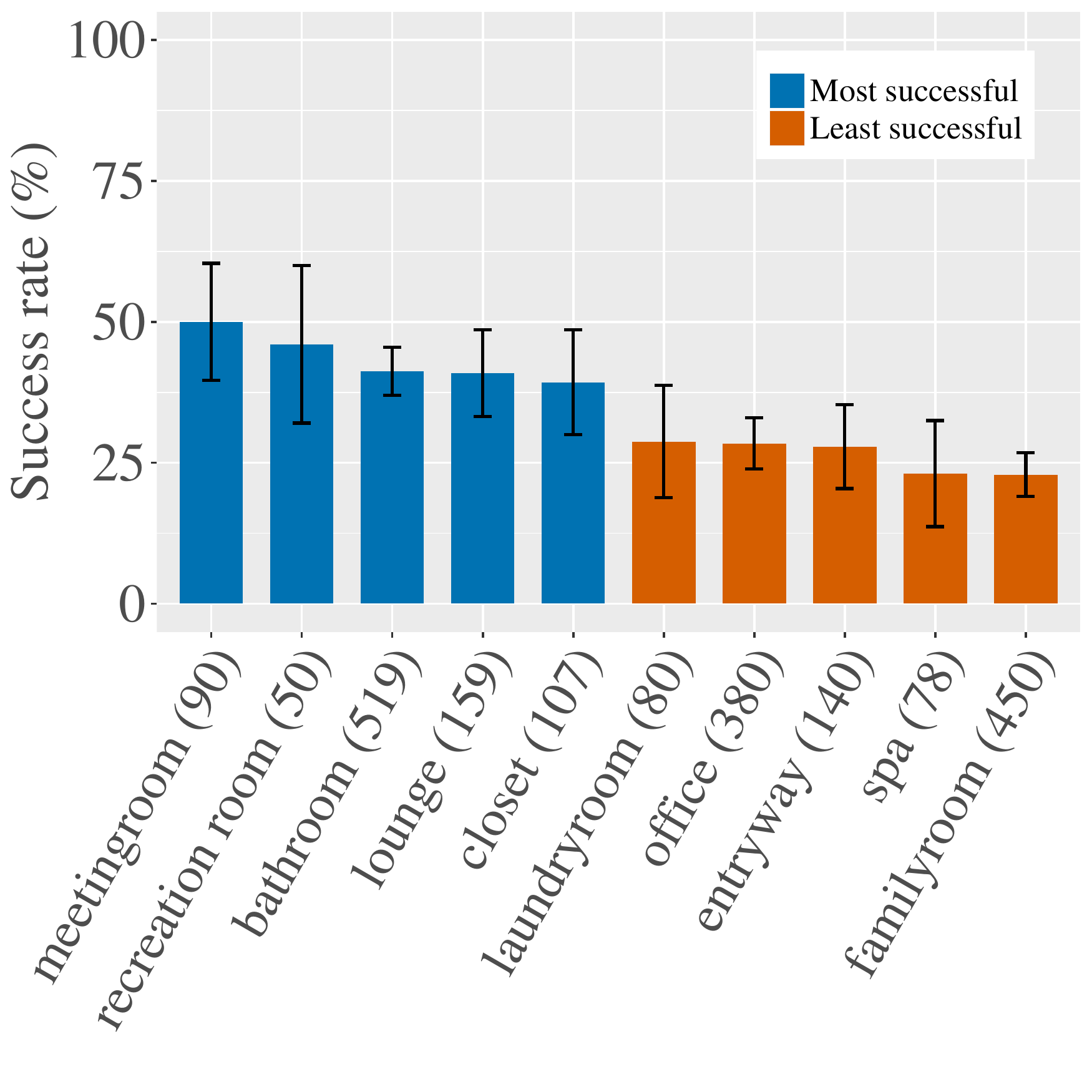}
        \caption{}
        \label{fig:success_room_unseen}
    \end{subfigure}
    \caption{Top five goal rooms with highest and lowest average success rates in (a) \textsc{Test seen} and (b) \textsc{Test unseen}. Numbers in parentheses are room frequencies. Only rooms appearing more than 50 times are included. Error bars are 95\% confidence intervals.}
    \label{fig:success_room}
\end{figure}

We analyze the behavior of an agent that is trained to learn a help-requesting policy (\textsc{Learned}) and is evaluated with a single random seed. 
This agent achieves a success rate of 52.0\% on \textsc{Test seen} and 34.5\% on \textsc{Test unseen}. 
Overall, the success rate of the agent degrades as the trajectory gets longer (Figure \ref{fig:success_len}). 
The agent tends to ask for help early (Figure \ref{fig:request_dist}), making more than half of its requests on the first 20\% steps. 
As time advances, the number of requests decreases. As expected, the agent tends to request help more early on \textsc{Test unseen} than on \textsc{Test seen}.
As shown in Figure \ref{fig:success_obj}, the most identifiable objects have distinct invariant features (e.g., sink, curtain), whereas the least identifiable objects greatly vary in shape and color (e.g., fireplace, stool, armchair). 
Mirrors are the easiest objects to detect in \textsc{Test Seen}, possibly because they are usually in small rooms (e.g., bathroom) and always reflect the camera used by the Matterport3D data collector. 
On \textsc{Test Unseen}, they are more difficult to find because walking to the containing-rooms is more challenging. 
Finding objects in bathrooms is less challenging because bathrooms are usually small and have similar layouts and locations among houses, whereas searching for objects in offices is difficult because the corresponding environments are usually workspaces that contain many similar-looking rooms (Figure \ref{fig:success_room}). 
Note that these rankings are subject to sampling biases; for example, they favor objects (or rooms) whose data points correspond to shorter paths.

\begin{table}[t!]
        \small
        \centering
        \begin{tabular}{lcc}
        \toprule
        \multicolumn{1}{c}{Teacher} & \textsc{Test Seen} & \textsc{Test Unseen}\\ \midrule
        \noask & 28.39  $\pm$ 0.00 & 6.36 $\pm$ 0.00 \\
        Rule (a), (e) ($\delta=2)$ & 30.36 $\pm$ 0.13 & 8.49 $\pm$ 0.08 \\
        Rule (a), (e) ($\delta=4)$ & 39.46 $\pm$ 0.05 & 8.78 $\pm$ 0.04 \\
        Rule (a), (e) ($\delta=8)$ & 30.71 $\pm$ 0.05 & 5.64 $\pm$ 0.00 \\
        Rule (b), (c), (d) & 51.89 $\pm$ 0.24 & 35.52 $\pm$ 0.29 \\
        All rules & 52.09 $\pm$ 0.13 & 34.50 $\pm$ 0.23 \\
        \bottomrule
       \end{tabular}
     \caption{Ablation study on the effectiveness of rules of the help-requesting teacher. All numbers are success rates (\%). See section 6.2 for specifications of the rules.}
     \label{tab:ablation_rule}
\end{table}

Table \ref{tab:ablation_rule} shows the effectiveness of different subsets of rules of the help-requesting teacher. 
Using only rules (b), (c), (d), which do not require learning because can be directly computed at test time without ground-truth information, is sufficient to obtain a success rate comparable to that of using all rules. 
Using rules (a) and (e), which require ground-truth information about the environment and the task, slightly improves the success rate over not requesting. 
Rules (a) and (e) are in fact very difficult to learn considering the small size of the Matterport3D dataset. 
Unfortunately, at the time this research was conducted, the Matterport3D simulator was one of the largest scale in the small pool of indoor simulators with real scenes. 
The effectiveness of rules (a) and (e) would be more visible in larger-scale environments like Gibson \cite{xiazamirhe2018gibsonenv} but it offered limited object annotation. 
Another reason that makes it challenging to learn rules (a) and (e) is the training-test condition mismatch. 
Near the end of training, the agent has memorized the training examples and rarely makes mistakes.
The agent is thus biased toward not requesting help and generalizes poorly to unseen examples. 
We nevertheless include rules (a) and (e) to illustrate that the help-requesting policy can be taught rules that cannot be executed at test time by a teacher that has access to ground-truth information. 
In general, imitating a help-requesting teacher allows us to easily transfer domain knowledge from humans to the agent without restriction on the knowledge and on information required to imitate it.

\end{document}